%% file: main_arxiv.tex
\PassOptionsToPackage{T2A,T1}{fontenc}
\PassOptionsToPackage{sort&compress}{natbib}
\documentclass{zlab}
\usepackage{tikz}
\usepackage[iclrfont]{gaze_arxiv}
\setcitestyle{numbers,square,comma}
\catcode`\_=8\relax
\RemoveFromHook{begindocument}[underscore]
\titlespacing*{\subsubsection}{0pc}{1ex plus1pt minus0.5pt}{0pt}

\usepackage{xspace}
\usepackage{multirow}

\usepackage{graphicx,amsmath,amssymb,hyperref}
\usepackage{subcaption}
\usepackage{longtable}
\usepackage{listings}
\usepackage{tcolorbox}
\tcbuselibrary{listings,skins,breakable,theorems}
\usepackage{adjustbox}
\usepackage{placeins}
\usepackage{wrapfig}
\usepackage{pdfpages}
\usepackage{etoc}

\DeclareUnicodeCharacter{043C}{{\fontencoding{T2A}\fontfamily{cmr}\selectfont\cyrm}}
\input{macros}
\newcommand{\projectleadmark}{\ast}
\newcommand{\corecontributormark}{\dagger}

\title{Learning to Persist in Visual Search}
\author{%
    \vspace{.2cm}
    \parbox{\textwidth}{\centering
        Gabriel Sarch$^{\projectleadmark,\corecontributormark}$ \quad
        Haoyang Wu$^{\corecontributormark}$ \quad
        Yipeng Zhang$^{\corecontributormark}$ \quad
        Linrong Cai$^{\corecontributormark}$ \\
        \vspace{.2em}\Authfont
        Katerina Fragkiadaki$^{\ddagger}$ \quad
        Danqi Chen \quad
        Zhuang Liu
    }
    \\
    \vspace{.35cm}
    {\normalfont\fontsize{11}{15}\selectfont
        Princeton University \quad
        $^{\ddagger}$Carnegie Mellon University}
    \vspace{.3cm}
}

\correspondingauthor{}

\renewcommand{\twocolumn}[1][]{\par#1\par}

\newcommand{\arxivWrapTable}[2][]{
    \begin{wraptable}[#1]{R}{0.50\textwidth}
        #2
    \end{wraptable}
}

\newcommand{\arxivSearchFigures}[2]{%
    \begin{figure}[t]
        \centering
        \captionsetup{position=bottom}
        \captionbox{#1}{\includegraphics[width=0.55\linewidth,trim=0 7.16bp 0 0,clip]{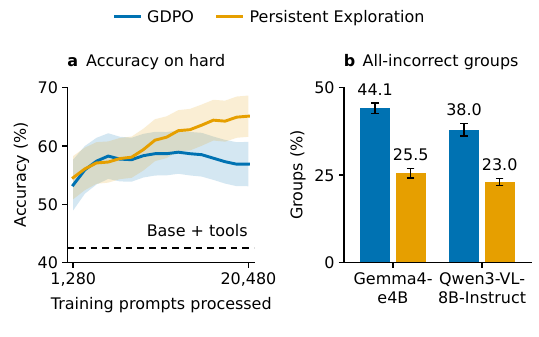}}\hfill
        \nextfloat
        \captionbox{#2}{\includegraphics[width=0.42\linewidth]{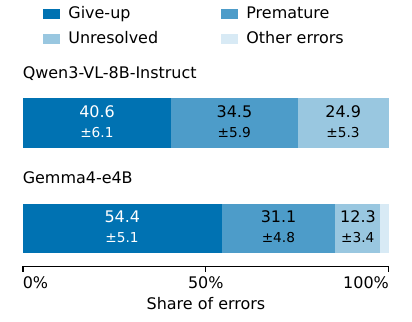}}
    \end{figure}
}

\begin{document}

\begingroup
\makeatletter
\let\raggedright\centering
\makeatother
\maketitle
\renewcommand{\thefootnote}{}
\footnotetext[0]{$^{\projectleadmark}$\,project lead\quad$^{\corecontributormark}$\,core contributor}
\endgroup

\vspace{-1.2cm}
\begin{center}
\small
\raisebox{-1.3pt}{\includegraphics[height=1.05em]{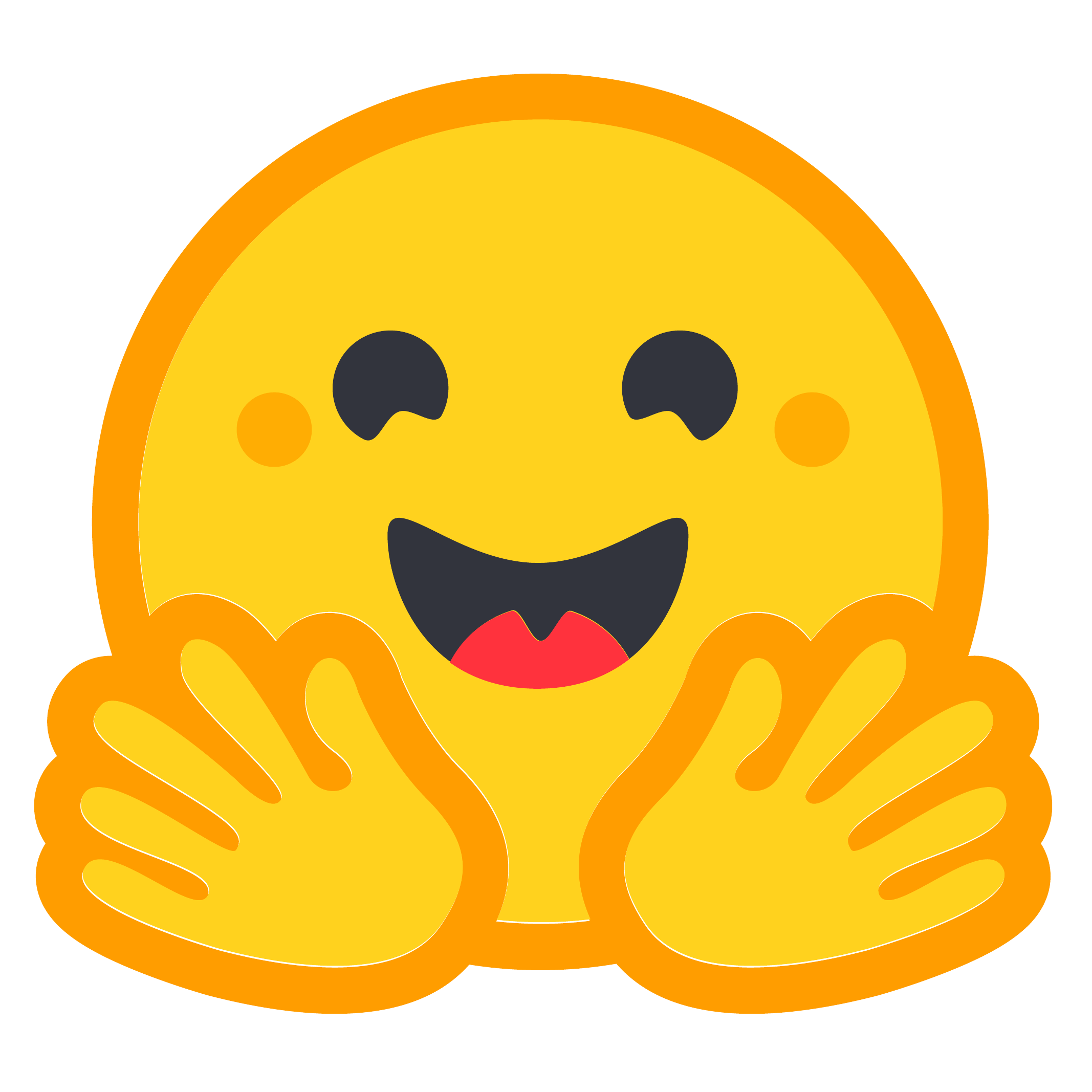}}\,
\href{https://huggingface.co/collections/gsarch/persistent-exploration}{\textcolor{verocolor}{\textbf{Models}}}
\qquad
\raisebox{-1.3pt}{\includegraphics[height=1.05em]{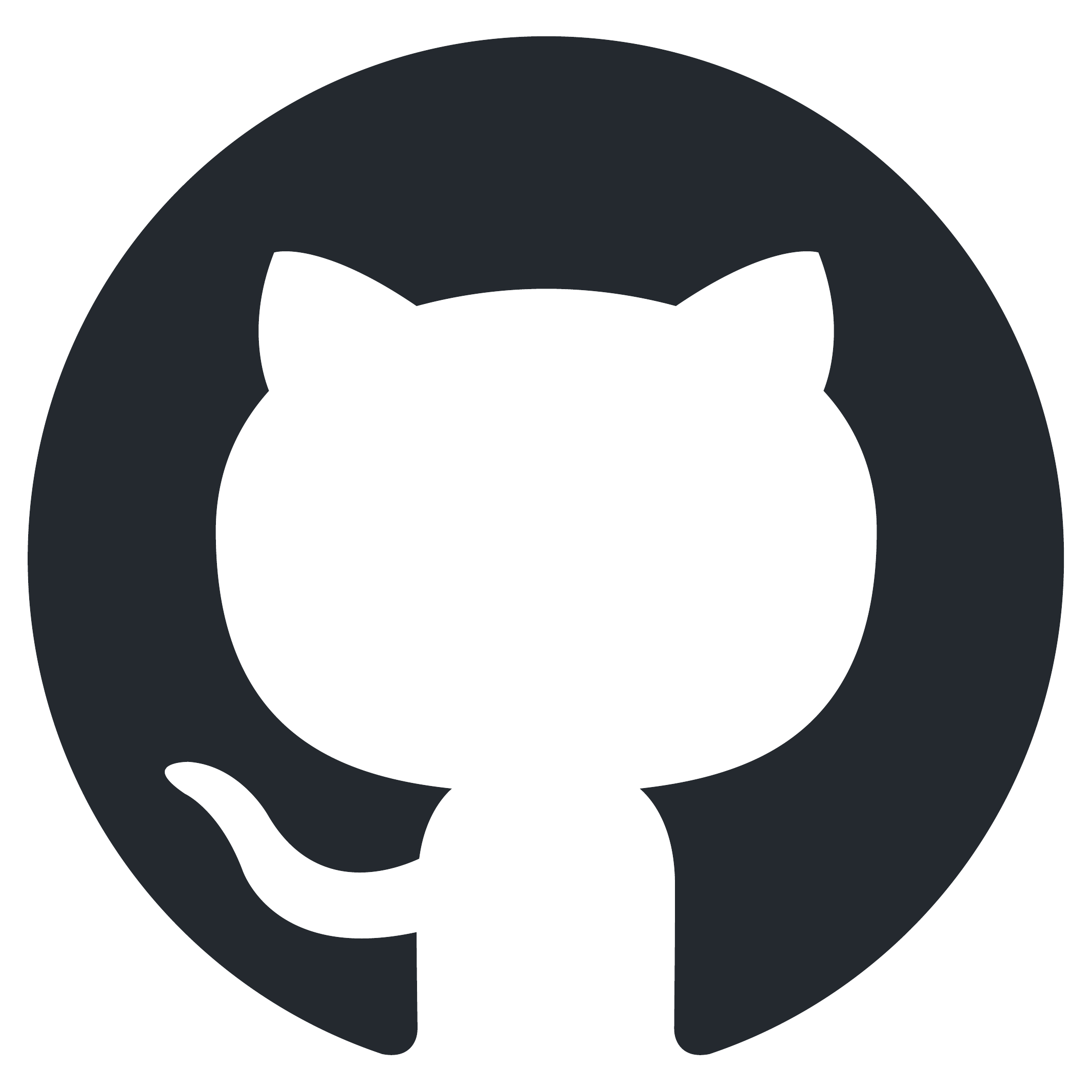}}\,
\href{https://github.com/Gabesarch/persistent-exploration}{\textcolor{verocolor}{\textbf{Code}}}
\qquad
\raisebox{-1.3pt}{\includegraphics[height=1.05em]{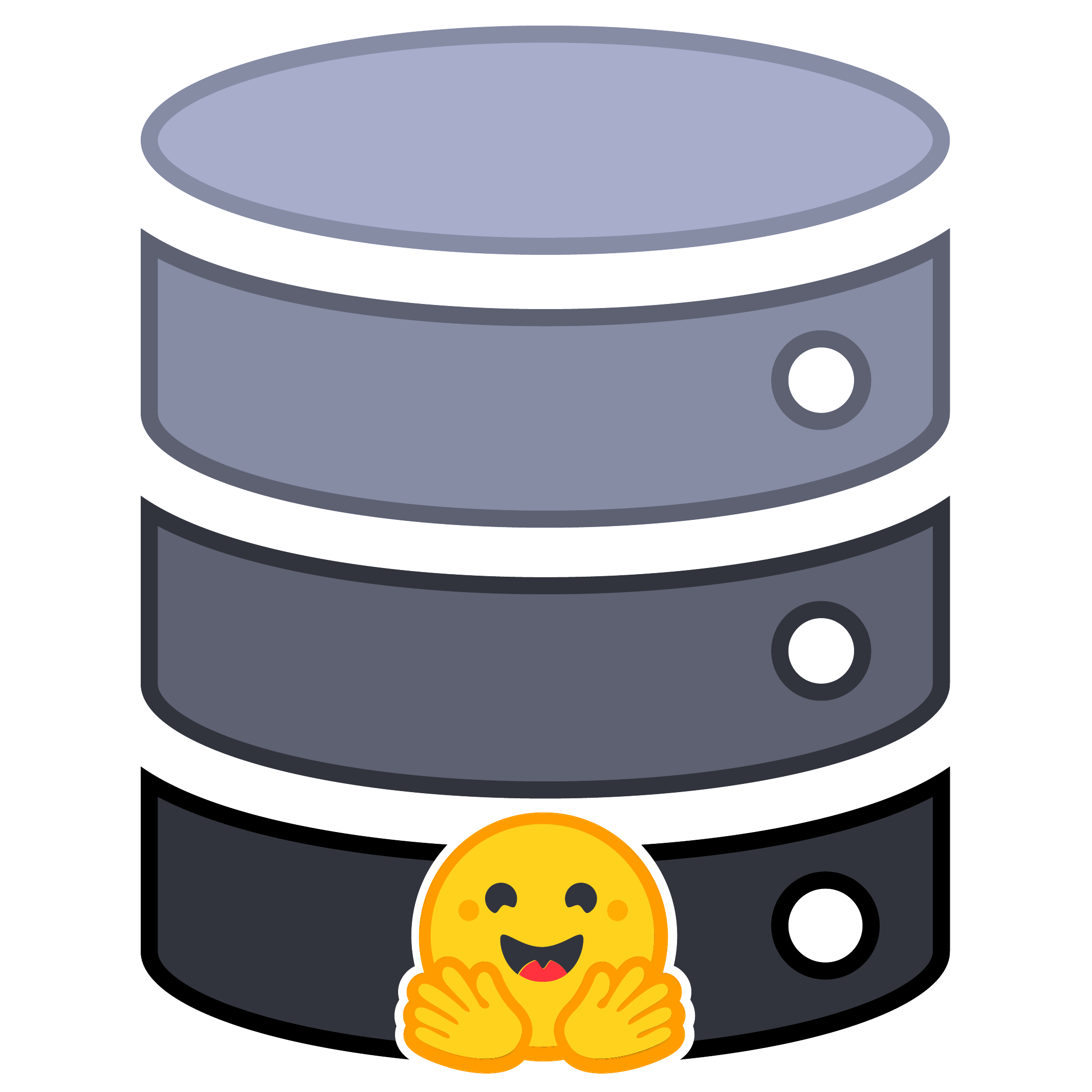}}\,
\href{https://huggingface.co/datasets/gsarch/Persistent-Exploration-RL-Data}{\textcolor{verocolor}{\textbf{Data}}}
\end{center}
\vspace{13pt}
\begin{abstractblock}
\vspace{-0.7em}
\input{sections/abstract}
\end{abstractblock}
\vspace{14pt}
\begingroup
\captionsetup{hypcap=false}
\begin{minipage}{\textwidth}
  \centering
  \captionsetup{type=figure}
  \includegraphics[width=0.93\textwidth]{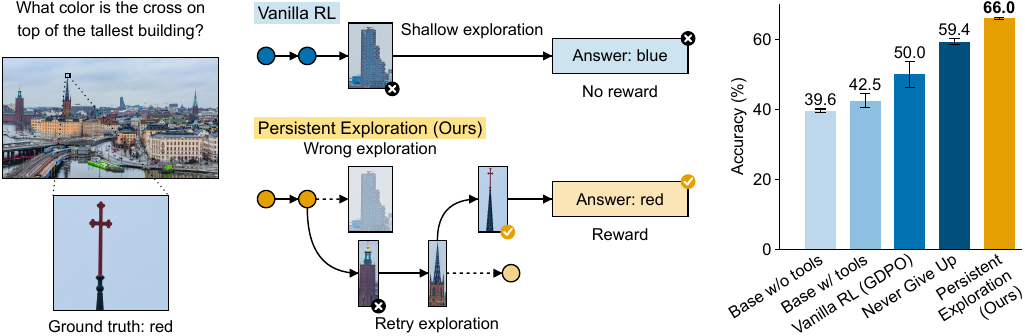}
  \caption{\textbf{Training vision-language models to persist after failure improves accuracy on hard visual search problems.} \textbf{(left)} Persistent exploration guides models to continue exploring upon search failure, enabling progress on hard problems where vanilla RL plateaus. \textbf{(right)} \method{} compared to the base model, vanilla RL (GDPO~\citep{liu2026gdpo}), and Never Give Up~\citep{noukhovitch2026learning} on the VisualProbe hard split (mean $\pm$ SEM on 3 eval runs). All trained variants start from Qwen3-VL-8B-Instruct, with Gemma4 in Table~\ref{tab:full_results}. }
  \label{fig:teaser}
\end{minipage}
\endgroup
\clearpage

\input{sections/introduction}

\input{sections/related_work}
\input{sections/method}
\input{sections/results}
\Needspace{7\baselineskip}
\input{sections/discussion_conclusion_long}

{
    \small
    \bibliographystyle{unsrtnat}
    \bibliography{references}
}

\clearpage
\input{sections/appendix}

\end{document}

%% file: macros.tex
\long\def\IfClassLoadedT#1#2{\IfClassLoadedTF{#1}{#2}{}}
\long\def\IfClassLoadedF#1#2{\IfClassLoadedTF{#1}{}{#2}}

\definecolor{findingborder}{HTML}{005B5B}
\definecolor{findingbackground}{HTML}{F2F2F2}
\newtcolorbox[auto counter]{findingbox}{
    enhanced,
    colback=findingbackground,
    colframe=findingborder,
    boxrule=0.65pt,
    arc=5pt,
    boxsep=0pt,
    left=11pt, right=11pt, top=7pt, bottom=7pt,
    before skip=8pt, after skip=8pt,
    shadow={2pt}{-2pt}{0pt}{black!12},
    fontupper=\normalsize,
    before upper={\textsf{\textbf{Finding~\thetcbcounter:}}\enspace}
}

\definecolor{verocolor}{HTML}{1976D2}
\definecolor{linkred}{HTML}{ED1C24}
\definecolor{linkmagenta}{HTML}{C2185B}

\definecolor{reviewcolor}{HTML}{D81B60}

\definecolor{chriscolor}{HTML}{2E7D32}

\newcommand{\method}{Persistent Exploration}
\newcommand{\rqnumber}[1]{\tikz[baseline=(rqnumber.base)]{\node[draw=black,circle,inner sep=0pt,minimum size=1em,line width=0.4pt] (rqnumber) {\small #1};}}

\definecolor{bestcellblue}{HTML}{EBF5FD}

\definecolor{tblred}{HTML}{FF8988}
\definecolor{tblorange}{HTML}{FECC81}
\definecolor{tblblue}{HTML}{6098FF}
\definecolor{tblgreen}{HTML}{77B25D}
\definecolor{tblpurple}{HTML}{B28CFF}
\definecolor{tblgray}{HTML}{8B8F87}

\definecolor{domGrounding}{HTML}{E3D6FF}
\definecolor{domChartOCR}{HTML}{FFD9D9}
\definecolor{domKnowledge}{HTML}{D4EACA}
\definecolor{domSpatial}{HTML}{FEE8C4}
\definecolor{domSTEM}{HTML}{D9E6FF}
\definecolor{domCaptioning}{HTML}{DDDEDA}
\definecolor{verosectionhl}{HTML}{D9E6FF}

\definecolor{promptblue}{HTML}{1976D2}
\definecolor{prompttext}{HTML}{0D2137}
\definecolor{promptborder}{HTML}{1976D2}
\definecolor{promptbg}{HTML}{F9F9F9}

\tcbset{
    promptstyle/.style={
        enhanced,
        breakable,
        listing only,
        colback=promptbg,
        boxrule=0.9pt,
        arc=3pt,
        left=10pt, right=10pt, top=8pt, bottom=8pt,
        listing options={
            language={},
            basicstyle=\fontsize{5}{6}\selectfont\sffamily,
            breaklines=true,
            breakatwhitespace=false,
            columns=fullflexible,
            keepspaces=true,
            showstringspaces=false,
            frame=none,
            upquote=true,
            xleftmargin=0pt,
            xrightmargin=0pt,
            aboveskip=0pt,
            belowskip=0pt,
        },
        fonttitle=\sffamily\fontseries{bx}\selectfont,
        toptitle=5pt,
        bottomtitle=5pt,
        lefttitle=10pt,
        boxsep=0pt,
        sharp corners
    }
}

\definecolor{promptcodebg}{HTML}{EFEFEF}

\tcbset{
    promptstylefmt/.style={
        breakable,
        colback=promptbg,
        boxrule=0.9pt,
        left=10pt, right=10pt, top=8pt, bottom=8pt,
        fontupper=\small\sffamily,
        fonttitle=\sffamily\fontseries{bx}\selectfont,
        toptitle=5pt,
        bottomtitle=5pt,
        lefttitle=10pt,
        boxsep=0pt,
        sharp corners
    }
}

\newcommand{\pbold}[1]{{\sffamily\bfseries #1}}
\newcommand{\pcode}[1]{{\small\ttfamily\colorbox{promptcodebg}{\strut #1}}}

\newcommand{\promptrule}{\par\vspace{4pt}\noindent\rule{\linewidth}{0.4pt}\par\vspace{4pt}}

\newenvironment{promptitemize}{%
    \begin{list}{\textbullet}{%
        \setlength{\topsep}{2pt}%
        \setlength{\itemsep}{1pt}%
        \setlength{\parsep}{0pt}%
        \setlength{\leftmargin}{1.5em}%
    }%
}{\end{list}}

\newcounter{promptenumi}
\newenvironment{promptenumerate}{%
    \begin{list}{\arabic{promptenumi}.}{%
        \usecounter{promptenumi}%
        \setlength{\topsep}{2pt}%
        \setlength{\itemsep}{1pt}%
        \setlength{\parsep}{0pt}%
        \setlength{\leftmargin}{2em}%
    }%
}{\end{list}}

\lstdefinestyle{promptcode}{
    basicstyle=\footnotesize\ttfamily,
    backgroundcolor=\color{promptcodebg},
    breaklines=true,
    columns=fullflexible,
    keepspaces=true,
    showstringspaces=false,
    frame=none,
    upquote=true,
    xleftmargin=4pt,
    xrightmargin=4pt,
    aboveskip=4pt,
    belowskip=4pt,
}

\lstdefinestyle{modelexampletrace}{
    basicstyle=\ttfamily\footnotesize,
    breaklines=true,
    mathescape=true,
}

%% file: sections/abstract.tex
\looseness=-1When learning to search for target objects in cluttered environments, vision-language models trained with standard reinforcement learning often make progress on easy problems but plateau on harder ones. We find that this plateau is due to models giving up or stopping prematurely after failed exploration, resulting in rollouts with no learning signal. We propose Persistent Exploration, a reinforcement learning method which guides models to continue exploring after unsuccessful attempts, using prior context to inform subsequent exploration. We update the policy to learn from the failed and recovered attempts, and to evaluate when further exploration is needed at test time. On a suite of challenging visual search tasks, encouraging persistent exploration leads to significant gains over standard reinforcement learning and methods that allocate additional independent rollouts to hard tasks. Persistent exploration training improves accuracy on hard VisualProbe problems from 50.0\% to 66.0\% over standard RL, and outperforms open general reasoners and specialized visual search models of similar size. Our trained models recover more often from failure and are less likely to give confident but incorrect answers prematurely. Our results demonstrate that persistence after failure is itself a powerful learning signal. 

%% file: sections/introduction.tex
\section{Introduction}
\label{sec:intro}

\looseness=-1Vision-language models (VLMs) are becoming increasingly expected to understand and process fine-grained details in visual content, such as text on a distant sign or small markers on a dense map. Answering such questions requires visual search, where a model must identify and reason about one or several small targets among many distractors in a high resolution image. VLMs typically encode images under a fixed visual token budget, such that fine-grained details are often lost when the image is encoded~\citep{wang2025retrieval,wei2026zooming,sarch2026grounded}. A common remedy to this is to allow the model to actively search by requesting high-resolution ``glimpses'' of parts of the visual scene, known as ``thinking with images''~\citep{zheng2026deepeyes,su2026pixel,sarch2026grounded,lai2026mini}. The problem then becomes a decision process about where to explore next in the visual scene and when the exploration is sufficient to answer.

Reinforcement learning has become a standard training technique during post-training of VLMs by deploying them in training environments and updating the model based on reward signals. While RL with zoom tools improves visual search accuracy in VLMs, the gains are largely concentrated in easier problems. For harder visual problems where targets become smaller relative to the image and the number of distractors grows, the model may need to explore many regions before becoming confident about an answer. On these harder problems, vanilla RL training plateaus (Section~\ref{sec:rl-plateaus}). This is consistent with prior work showing RL enables models to make progress on easy tasks near the edge of competence, while harder tasks that are well beyond the capabilities of the model often see little progress throughout the course of training~\citep{zhang2512interplay,qu2026pope}. In sparse reward settings like visual search, a model which consistently does not find the answer receives little learning signal for hard problems.

\looseness=-1We find that a major reason for this plateau on hard visual search tasks is that models fail to persist in exploration after unsuccessful attempts. When given a hard visual search task, we show that models equipped with visual tools explore shallowly, giving up after not finding the answer quickly, and answering prematurely despite not being confident (Section~\ref{sec:exploration-analysis}). This behavior impacts their progress during RL training as models tend to give up or answer with limited evidence, leading to RL episodes with no reward and no learning signal. \citet{noukhovitch2026learning} have proposed learning hard problems by allocating more rollouts to unsolved problems. However, by simply running more rollouts from scratch, each attempt cannot observe the previous failures and thus does not learn to adjust their strategy based on past attempts (Section~\ref{sec:results-rq2}). Other works such as \citet{lai2026mini} and \citet{sarch2026grounded} try to address this by shaping rewards, or forcing longer trajectories at test time~\citep{muennighoff2025s1}. However, we also show that these solutions show limited improvements (Section~\ref{sec:results}).

We find that training models to persist after failed attempts allows them to recover from exploration limitations and improve on hard problems. We propose \method{}, a method which addresses shallow visual tool use exploration during RL by guiding models to explore alternative solutions after failing to reach a correct answer. Instead of treating unsuccessful attempts as independent episodes, we construct a persistent exploration episode by continuing exploration after failure and conditioning subsequent attempts on the accumulated context from prior attempts. When a rollout fails, we remove the failed answer context and append a fixed prefix encouraging the model to continue to explore and forcing additional tool calls before the model is able to answer again. This simple modification enables models to recover from an average of 20\% of failed attempts during RL rollouts. We then train on both the recovered rollouts and failed attempts to internalize such behavior, and add an auxiliary evaluator loss that trains the model to evaluate whether its answers have succeeded or failed.

\looseness=-1We train on VisualProbe~\citep{lai2026mini} and evaluate on a suite of visual search benchmarks including VisualProbe-test, V*, PinpointBench, and O3Bench. On the hard split of VisualProbe, \method{} improves over GDPO from 50.0\% to 66.0\% for Qwen3-VL-8B and from 27.4\% to 40.6\% for Gemma4-E4B, demonstrating how learning persistent exploration can overcome the exploration bottleneck of vanilla on-policy exploration. \method{} outperforms recent methods that retry unsolved problems from scratch~\citep{noukhovitch2026learning}, showing how learning from the failed context is important.
Trained on Qwen3-VL-8B, \method{} surpasses GPT-6-luna and Sonnet 5 on VisualProbe and PinpointBench, while outperforming visual reasoners of similar size.
A behavioral analysis reveals that \method{} promotes more systematic visual search, and trained models revise their exploration strategy after failure and stop less often before exploring sufficient evidence. Our results suggest that persistence after failure is a powerful learning signal for visual search agents.

%% file: sections/related_work.tex
\section{Related Work}
\label{sec:related-work}

\noindent\textbf{Tool-Augmented Visual Search.}
Visual question answering requiring small targets in high resolution is difficult for models due to visual encoding blurring the small details and large numbers of visual tokens overwhelming context. This has motivated the use of a zoom in tool, where models can selectively request higher resolution crops of regions of the image, known as ``thinking with images''~\citep{zheng2026deepeyes,su2026pixel,sarch2026grounded,lai2026mini}. DeepEyes~\citep{zheng2026deepeyes} showed that reinforcement learning can teach a model to request zoomed crops without additional supervised tool traces. Pixel Reasoner~\citep{su2026pixel} added a curiosity reward to counter the model's tendency to avoid zoom, while ViGoRL~\citep{sarch2026grounded} adds an exploration bonus reward. Mini-o3~\citep{lai2026mini} avoids penalizing trajectories that reach the training turn limit. These models rely on the ability of the initial model to explore effectively and obtain reward. 
Retrieval-Augmented Perception~\citep{wang2025retrieval}
retrieves relevant image regions, whereas Zooming without Zooming~\citep{wei2026zooming} distills region-level supervision into a single-pass model. Our work explores how to train models to persist in their exploration when a single pass or extra rewards are not enough to make progress on the hardest tasks.

\noindent\textbf{Reinforcement Learning in Visual Environments.}
Reinforcement learning provides a scalable framework to post-train vision-language models from outcome-based feedback without dense expert supervision. While recent work demonstrates that scaling multi-task RL across broad visual domains substantially boosts general visual reasoning~\citep{sarch2026vero}, an important branch of research applies RL to interactive visual environments where agents can manipulate images via callable tools~\citep{zheng2026deepeyes,lai2026mini}. In these settings, RL is typically utilized to optimize tool strategies, multi-turn visual search, or adaptive tool composition~\citep{dong2026toolsrl,wu2026vctooler}. Rather than allowing models to terminate prematurely, our approach guides agents to persist in exploration upon failure.

\looseness=-1\noindent\textbf{Learning Hard Problems with Reinforcement Learning.}
Reinforcement learning on hard, sparse-reward problems has historically paired
search with verification. AlphaZero~\citep{silver2018general} combines policy
and value networks with tree search in board games. Without explicit search, group-based RL mostly sharpens solutions the base model can already sample, stalling on hard problems when every rollout fails and group advantages drop to zero~\citep{yue2025does,zhang2512interplay}. Recent methods have explored addressing all-wrong groups by filtering them out~\citep{yu2025dapo},
sampling extra rollouts from scratch~\citep{noukhovitch2026learning}, or
starting from partial oracle solutions~\citep{qu2026pope}. Our approach keeps the prior zoom history in context so retries utilize that information, without needing oracle
supervision beyond the final answer. Our work connects to methods that continue from visited states or failed attempts, including Go-Explore~\citep{ecoffet2021first}, multi-turn
self-correction~\citep{kumar2025training,qu2025optimizing}, and test-time budget forcing~\citep{muennighoff2025s1}. Rather than resetting a simulator or revising text over a fixed input, we encourage exploration after failure, and penalize the failed answer branches. Finally, instead of forcing extra turns at test time, the model learns to decide when to keep searching.

\noindent\textbf{Cross-Episode Meta-Reinforcement Learning.}
Our work is related to meta-reinforcement learning, where an agent builds on prior interactions to improve their behavior. In cross-episode meta-RL, interaction histories are retained across rollouts, allowing the policy to update their parameters conditioned on previously collected experience. Recent approaches, such as Efficient Cross-Episode Meta-RL~\citep{shala2025efficient} and ORBIT~\citep{lin2026scaling}, show that training on trajectories that combine multiple episodes enables agents to learn improved exploration and adaptation strategies in context. Our work uses a similar strategy for visual search, where instead of independent attempts, we construct experience in which subsequent attempts build on prior failed context. Unlike cross-episode meta-RL, which trains agents across repeated environment episodes, our method intervenes at unsuccessful answer terminations within a single visual search task.

%% file: sections/method.tex
\section{Problem Setup}\label{sec:prelim}

Our goal is to learn a visual reasoner, $\pi_\theta$, that uses textual reasoning and tools to help it solve visual tasks. We consider visual search where the model receives a high resolution image, $I$, and a question, $q$, and must output a predicted answer, $a$, that matches the ground truth answer $a^*$. The difficulty in visual search lies in identifying a target that is a small proportion of the visual input and is among many distracting objects. 

We provide the visual reasoner with a zoom tool that takes in a bounding box $Z_t = [x_1, y_1, x_2, y_2]$ and returns a zoomed in crop $C_t$ of the initial image $I$. The model interleaves textual reasoning $T_t$ with zooming, following $T_t \rightarrow Z_t \rightarrow C_t \rightarrow T_{t+1}$ until it outputs an answer $a$.

We consider a training dataset $\mathcal{D}$ of $(I, q, a^*)$ triples and train $\pi_\theta$ with reinforcement learning from outcome rewards. For each $(I, q)$, the model samples a trajectory $\tau$ of interleaved reasoning, zooms, and crops that ends in an answer $a$. The trajectory receives a reward $R(\tau) = \sum_j w_j\, r_j(\tau)$ that combines an accuracy reward $r_{\text{acc}} \in \{0, 0.5, 1\}$, indicating whether $a$ matches $a^*$, an efficiency reward $r_{\text{eff}}$ for solving the problem in fewer turns, and a format reward $r_{\text{fmt}}$ for following the reasoning and tool-call format. 
We train $\pi_\theta$ to maximize $J(\theta) = \mathbb{E}_{(I, q, a^*) \sim \mathcal{D}}\,
    \mathbb{E}_{\tau \sim \pi_\theta(\cdot \mid I, q)} \big[ R(\tau) \big]$.
During training, we sample $k$ rollouts for each input and compute GDPO~\citep{liu2026gdpo} advantages as $A_i = \operatorname{Norm}_{\mathcal{B}}\!\left(\sum_j w_j \frac{r_j(\tau_i)-\mu_j}{\sigma_j+\epsilon}\right)$, where $i$ indexes rollouts, $\mu_j$ and $\sigma_j$ are the mean and standard deviation of reward component $j$ across the $k$ rollouts, $\epsilon > 0$ is a numerical stabilizer, and $\operatorname{Norm}_{\mathcal{B}}$ denotes normalization across the training batch $\mathcal{B}$. We update the model using GSPO~\citep{zheng2025groupsequencepolicyoptimization} policy gradient updates.
We provide further details of the RL training in the Appendix~\ref{app:training}.

\newcommand{\trainingComparisonCaption}{\textbf{Persistent exploration improves on hard visual search problems where vanilla RL plateaus}. \textbf{a.} Accuracy on VisualProbe hard set over training steps. Vanilla RL makes some progress over the base model but then quickly plateaus. \textbf{b.} The average number of rollout groups where all trajectories are incorrect across the first 50 steps of training.}
\newcommand{\failureModesCaption}{\textbf{Models fail to persist after exploring only a few possibilities.} Behaviors are categorized into one of four error modes. Values show mean percentages \textpm{} 95\% CI across samples.}
    \arxivSearchFigures
        {\trainingComparisonCaption\label{fig:training_comparison}}
        {\failureModesCaption\label{fig:base-model-error-breakdown}}

\subsection{Vanilla RL plateaus on hard visual search}
\label{sec:rl-plateaus}

We begin our analysis by examining a vanilla RL setting where we sample on-policy trajectories and apply GDPO updates, as described in Section~\ref{sec:prelim}. We use the VisualProbe~\citep{lai2026mini} training set filtered for examples that a text-only reasoning model, Vero-Qwen3i-8B~\citep{sarch2026vero}, cannot solve in 16 attempts, yielding 2,042 samples. We test on the held out VisualProbe set~\citep{lai2026mini}, which is split into easy, medium, and hard difficulty problems based on the size of the input image, number of distracting elements, and the size of the target relative to the total image size. We report results for hard problems in Table~\ref{tab:full_results} and Figure~\ref{fig:training_comparison}.

\looseness=-1The results reveal that while vanilla RL improves above the base model (50.0\% vs. 42.5\% for the base model on hard problems), the model quickly plateaus during RL training. On hard problems, accuracy improves early but shows no sustained gains with further training (Figure~\ref{fig:training_comparison}a; 53.8\% at step 10 versus 50.0\% at step 200). We attribute this plateau to not consistently obtaining reward on hard problems during training. In particular, we find a large portion of rollout groups receive no accuracy reward, providing no learning signal for the model (Figure~\ref{fig:training_comparison}b; 44.1\% of groups were all incorrect during the first 50 steps). We show in Section~\ref{sec:results-rq2} that simply adding more rollouts to harder problems~\citep{noukhovitch2026learning} or masking overlong turns~\citep{lai2026mini} leads to limited ability to overcome this plateau. Thus, using vanilla on-policy rollouts, the models fail to explore sufficiently on difficult visual search problems, and further RL training alone does not resolve this issue. 

\begin{findingbox}
Vanilla RL plateaus on hard visual search problems, with a large proportion obtaining no reward.
\end{findingbox}

\subsection{Models terminate exploration prematurely}
\label{sec:exploration-analysis}

We next examine the models from the previous Section~\ref{sec:rl-plateaus} to identify which aspects of their exploration behavior limit their ability to solve hard visual search problems. For each model failure, we categorize the behavior of the model into one of four failure modes using GPT-6.1-xhigh. Full details can be found in the Appendix~\ref{app:example-trajectories}. We validate the model ratings with a human annotator ($N=50$ samples; $96\%$ agreement with model). 

\looseness=-1We show the results of the failure analysis in Figure~\ref{fig:base-model-error-breakdown}. Across both Qwen3-VL-8B-Instruct and Gemma4-E4B, we find that most failures fall into the giving up or premature answer categories (75.1\% and 85.4\% of failures, respectively). Figure~\ref{fig:behavior-taxonomy} shows that premature stopping persists after GDPO training. These results suggest that the models lack the ability to persist in their exploration after a failed attempt, leading to low performance on difficult tasks where models may need to search through many potential solutions before finding the answer.

\begin{findingbox}
Models often stop exploring prematurely, leading to limited ability to solve hard problems.
\end{findingbox}

\section{Teaching Models to be Persistent Explorers}
\label{sec:our_method}

In Sections~\ref{sec:rl-plateaus}--\ref{sec:exploration-analysis}, we showed how standard RL on existing VLMs plateaus on hard visual search tasks and models often fail to explore sufficiently after a failed attempt. Building on these findings, we propose a method to encourage persistent exploration when failure occurs. While prior work has proposed using guidance or partial solutions obtained from an oracle or strong model~\citep{qu2026pope,zhang2026bread,zhang2026stephint,wang2026beacon}, we wish to improve exploration without such privileged information.

\looseness=-1\noindent\textbf{Proposed approach.} Our key insight is that encouraging models to persist after failure can uncover solutions that their initial exploration missed. While prior work has explored sampling additional fresh attempts from scratch for the hardest problems (e.g., \citet{noukhovitch2026learning}), we argue that persistent exploration should use the prior failed context to diagnose its failed attempt and decide the best alternative strategy to try next. Our proposed method makes three modifications to vanilla on-policy reinforcement learning: (1) guiding persistent exploration during rollouts (Section~\ref{sec:methods_persistent_explore}), (2) a policy gradient loss for failed attempts (Section~\ref{sec:methods_loss}), and (3) learning when to retry (Section~\ref{sec:learning_retry}).

\begin{figure*}[!t]
    \centering
        \includegraphics[width=\textwidth]{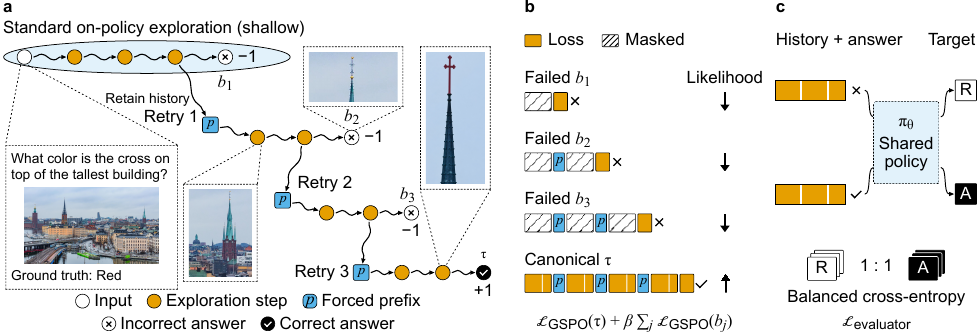}
    \caption{\textbf{\method{} guides models to retry after failure and learns from the canonical and failed attempts.} \textbf{a.} When an answer is incorrect during training, we remove the last attempt of reasoning and answer while keeping the preceding history, append a prefix to encourage exploration, and force another tool call. We continue this process until the model succeeds or hits a predefined budget. \textbf{b.} Policy gradient updates learn from the canonical trajectories and the failed attempts with masked shared context. Prefixes get masked for both. \textbf{c.} We add an auxiliary cross entropy loss on the same policy to predict whether an answer should be accepted (A) or retried (R).}
    \label{fig:methods_figure}
\end{figure*}

\subsection{Guiding persistent exploration during rollouts}\label{sec:methods_persistent_explore}
A key limitation of standard on-policy RL is that the model itself decides when to stop. However, when the problem is difficult and the model fails to explore sufficiently, this can lead to little learning signal. Instead, we intervene when the model is incorrect, and force it to keep exploring until it finds the answers or exhausts its exploration budget.

\looseness=-1 We guide persistent exploration as shown in Figure~\ref{fig:methods_figure}. We rollout the model until it outputs an answer $a$, and evaluate the answer using the accuracy reward $r_{\text{acc}}$. If the answer is correct, the episode terminates. If the answer is incorrect, we remove the last attempt of reasoning and answer, while keeping the preceding exploration history \(\tau_{<t}\) in context. We then append a fixed prefix $p$ to the beginning of the textual reasoning $T_t$ in the next assistant turn that encourages the model to continue exploring (\textit{``Let’s see. I still don’t have decisive evidence, so I should inspect a different region...''}). To prevent the model from providing an immediate answer in the next turn, we force the next action to be a tool call by constraining the decoding to output the \texttt{<tool\_call>} tag after the reasoning block. We repeat this procedure until the model gives a correct answer or reaches a turn or token budget. This process produces the canonical trajectory $\tau$ and a set of trajectories from the failed attempt $b_1,\ldots,b_m$.

We repeat this intervention whenever the model produces an incorrect answer and continue the exploration from the accumulated reasoning-tool-observation history. Unlike adding independent restarts for harder problems~\citep{noukhovitch2026learning}, subsequent exploration builds upon prior exploration context from the same episode. Thus, this intervention extends computation for harder problems by allowing the model to observe prior exploration context and encouraging the model to explore alternative solutions that it has not tried previously. 

This process of guiding persistent exploration after failure leads to a recovery rate (the fraction of initial attempts that fail and then eventually succeed after retrying) of 18.4--28.6\% (Appendix~\ref{app:tool-call-budgets-and-retry-training}). Consequently, all-incorrect rollout groups drop by approximately 40\% compared to vanilla on-policy exploration (Figure~\ref{fig:training_comparison}; Qwen3-VL-8B, 38.0\% vanilla RL vs. 23.0\% persistent exploration; Gemma4-E4B, 44.1\% vanilla RL vs. 25.5\% persistent exploration).

\begin{findingbox}
Persistent exploration recovers from initial failures and reduces all-incorrect groups by 40\%.
\end{findingbox}

\subsection{A policy gradient loss for failed attempts}\label{sec:methods_loss}

\noindent\textbf{Experience construction.} Given the persistent exploration experience from Section~\ref{sec:methods_persistent_explore}, we now wish to update the model to internalize the exploration behavior that led to success and downweight failed attempts. If we only update the model based on the canonical trajectories $\tau$, the failed early answers $b_1,\ldots,b_m$ would never be penalized. We thus keep each failed trajectory as an extra training sample that shares history in $\tau$. We apply loss only to the failed part after the branch point (i.e., the early wrong answer) while masking the shared context. We keep the prefixes in the context during training and mask them. Since these prefixes are off-policy, unmasking them to include them in the loss leads to unstable training and lower performance (see Appendix~\ref{app:tool-call-budgets-and-retry-training}).

\noindent\textbf{Advantage and loss computation.} We compute the advantage for each sample using the GDPO advantage calculation (Section~\ref{sec:prelim}). We compute all group means and standard deviations using only the $k$ canonical trajectories $\{\tau_i\}_{i=1}^{k}$ within the group, and then score all attempts, both canonical and failed, against them. Failed attempts thus generally get a negative advantage when the group has a mix of both correct and incorrect answers. Our loss then combines the GSPO update of the canonical trajectories $\{\tau\}$ and the failed trajectories $b_1,\ldots,b_m$, weighted by $\beta$:
\begin{equation*}
\setlength{\abovedisplayskip}{4pt}
\setlength{\belowdisplayskip}{4pt}
\setlength{\abovedisplayshortskip}{0pt}
\setlength{\belowdisplayshortskip}{0pt}
\mathcal{L} = \mathcal{L}_{\text{GSPO}}(\tau) + \beta \sum_j \mathcal{L}_{\text{GSPO}}(b_j)
\end{equation*}
The canonical trajectory loss $\mathcal{L}_{\text{GSPO}}(\tau)$ thus updates the model towards the persistent exploration behavior we desire, while the failed attempt loss $\mathcal{L}_{\text{GSPO}}(b_j)$ updates the model against incorrect and premature answers.

\subsection{Learning when to retry}\label{sec:learning_retry}

During training, retry happens via a check against a ground truth answer. However, at test time, the model must verify for itself whether to retry or not. In addition to the policy gradient losses, we add an auxiliary loss for the model to judge whether to accept an answer or continue exploring. For each incorrect and correct answer provided by the model, we append a fixed accept/retry question (``Should the answer be accepted or retried?''). We train with cross-entropy ($\mathcal{L}_{\text{evaluator}}$) on a single A/R output token and balance A/R classes across prompts. The full objective becomes:
\begin{equation*}
\setlength{\abovedisplayskip}{4pt}
\setlength{\belowdisplayskip}{4pt}
\setlength{\abovedisplayshortskip}{0pt}
\setlength{\belowdisplayshortskip}{0pt}
\mathcal{L} = \mathcal{L}_{\text{GSPO}}(\tau) + \beta \sum_j \mathcal{L}_{\text{GSPO}}(b_j) + \lambda\, \mathcal{L}_{\text{evaluator}}
\end{equation*}
During inference, we append the retry question after answers and force additional exploration if $P(R)>0.5$.

%% file: sections/results.tex
\newcommand{\ablationTable}[1][3pt]{
    \centering
    \small
    \setlength{\tabcolsep}{#1}
    \renewcommand{\arraystretch}{1.12}
    \begin{adjustbox}{max width=\columnwidth}
    \begin{tabular}{@{}lrrrr@{}}
        \toprule
        \multirow{2}{*}{Condition} & \multicolumn{4}{c}{VisualProbe} \\
        \cmidrule(lr){2-5}
        & Easy & Medium & Hard & Overall \\
        \midrule

\method{} & \textbf{78.7} & \textbf{67.5} & \textbf{66.0} & \textbf{70.3} \\
w/o $\mathcal{L}_{\text{evaluator}}$ & 73.0 & 64.2 & 64.2 & 66.6 \\
w/o $\mathcal{L}_{\text{GSPO}}(b_j)$ & 73.0 & 66.0 & 65.1 & 67.8 \\
        \bottomrule
    \end{tabular}
    \end{adjustbox}
    \caption{\looseness=-1\textbf{The policy gradient loss on the failed attempts and evaluator loss both contribute to performance.} Reported is VisualProbe accuracy (\%). All ablations train on Qwen3-VL-8B-Instruct.}
    \label{tab:visual-probe-ablations}
}

\newcommand{\behaviorFigure}{
    \centering
    \adjustbox{max width=\linewidth}{\includegraphics{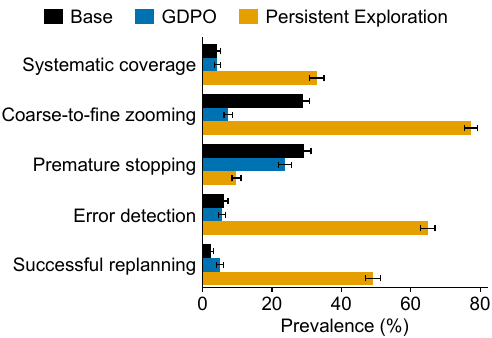}}
    \caption{\textbf{Persistent exploration exhibits more sustained search and ability to alter its search strategy upon failure.} Behavior presence rates for the base model (Qwen3-VL-8B-Instruct), vanilla RL, and \method{} on VisualProbe test trajectories. Full behavior definitions are in the Appendix~\ref{app:example-trajectories}.}
    \label{fig:behavior-taxonomy}
}

\newcommand{\qualitativeFigure}{
    \centering
    \includegraphics[width=\textwidth]{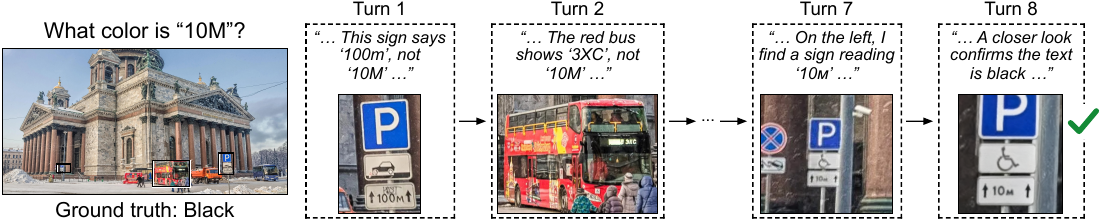}
    \caption{\textbf{Persistent exploration recognizes unsuccessful exploration and adapts its search strategy to confidently find the answer.} \method{} identifies a related but incorrect target sign (``100m'' sign) and searches another region to find the correct one. The model then progressively zooms into the target region to verify it has the correct target.}
    \label{fig:qualitative_example}
}

\section{Results}
\label{sec:results}

Our experiments aim to understand how persistent exploration training affects the visual search performance in vision-language models. In particular, we seek to understand the following research questions:\\*[6pt]
\rqnumber{1}~\textbf{RQ1:} How much does persistent exploration training help when evaluated on visual search tasks? (Section~\ref{sec:results-rq1})\\*[6pt]
\rqnumber{2}~\textbf{RQ2:} How important is learning from failed attempts for persistent exploration? (Section~\ref{sec:results-rq2})\\*[6pt]
\rqnumber{3}~\textbf{RQ3:} How does persistent exploration training change the behavior of the model? (Section~\ref{sec:behavior-analysis})\par

\subsection{Experimental setup}

\looseness=-1We investigate how persistent exploration training compares to strong reinforcement learning methods and strong general reasoning and visual search baselines on a variety of visual search benchmarks. We use the following experimental setup.

\ifdefined\arxivWrapTable
    \newenvironment{fullresultstable}
        {\wraptable[36]{R}{0.50\textwidth}}{\endwraptable}
\else
    \newenvironment{fullresultstable}
        {\suppressfloats[t]\table[t]}{\endtable}
\fi
\begin{fullresultstable}
    \centering
    \fontsize{8.5}{10}\selectfont
    \setlength{\tabcolsep}{0.35pt}
    \renewcommand{\arraystretch}{1.12}
    \begin{tabular}{@{}l*{7}{>{\centering\arraybackslash}p{21.65pt}}@{}}
        \toprule
        \multicolumn{1}{@{}c}{\multirow[c]{2}{*}[-2pt]{Condition}} & \multicolumn{4}{c}{VisualProbe} & \multirow[b]{2}{21.65pt}[-\dimexpr\aboverulesep+\cmidrulewidth+\belowrulesep\relax]{\centering\makebox[21.65pt][c]{\shortstack{Pinpoint\\Bench}}} & \multirow[b]{2}{21.65pt}[-\dimexpr\aboverulesep+\cmidrulewidth+\belowrulesep\relax]{\centering V\rlap{\textsuperscript{*}}} & \multirow[b]{2}{21.65pt}[-\dimexpr\aboverulesep+\cmidrulewidth+\belowrulesep\relax]{\centering\makebox[21.65pt][c]{\shortstack{O3\\Bench}}} \\
        \cmidrule(lr){2-5}
        & \makebox[21.65pt][c]{Easy} & \makebox[21.65pt][c]{Med.} & \makebox[21.65pt][c]{Hard} & \makebox[21.65pt][c]{All} & & & \\
        \midrule
        \multicolumn{8}{@{}l@{}}{\emph{Closed models}} \\
GPT-6-luna & 67.4 & 48.9 & 48.1 & 53.8 & 54.3 & 88.5 & 75.7 \\
GPT-6-luna + tools & 61.0 & 52.6 & 47.2 & 53.8 & 55.0 & 88.0 & 79.4 \\
Sonnet 5 & 53.9 & 34.3 & 26.4 & 38.1 & 34.4 & 77.0 & 64.1 \\
Sonnet 5 + native tools & 58.2 & 47.4 & 29.2 & 46.6 & 49.4 & 88.0 & 73.6 \\
Sonnet 5 + zoom tool & 72.3 & 64.9 & 60.4 & 66.0 & 61.4 & 92.1 & 84.9 \\
        \midrule
        \multicolumn{8}{@{}l@{}}{\emph{Open general and specialized search models}} \\
Qwen3.5-9B~\citep{qwen3.5} & 58.9 & 42.9 & 45.3 & 47.8 & 48.3 & 90.1 & 47.8 \\
Qwen3.5-9B + tools & 70.9 & 50.4 & 53.8 & 56.7 & 48.7 & 83.3 & 41.7 \\
Vero-Qwen3i-8B~\citep{sarch2026vero} & 62.4 & 41.0 & 47.2 & 48.2 & 44.1 & 87.4 & 49.0 \\
Mini-o3-7B~\citep{lai2026mini} & 67.0 & 50.4 & 48.0 & 54.5 & 44.3 & 88.2 & 29.9 \\
DeepEyes-7B~\citep{zheng2026deepeyes} & 60.1 & 29.8 & 35.1 & 39.2 & 39.7 & 83.3 & 28.4 \\
PixelEyes-8B~\citep{gong2026pixeleyes} & 71.6 & 55.2 & 59.4 & 60.6 & 55.2 & 94.2 & 34.2 \\
        \midrule
        \multicolumn{8}{@{}l@{}}{\emph{RL comparisons (Qwen3-VL-8B)}} \\
Qwen3-VL-8B~\citep{bai2025qwen3} & 61.7 & 41.0 & 39.6 & 46.4 & 40.7 & 86.9 & 46.7 \\
Qwen3-VL-8B + tools & 61.0 & 50.4 & 42.5 & 51.7 & 47.6 & 89.0 & 41.2 \\
GDPO~\citep{liu2026gdpo} & 73.8 & 61.9 & 50.0 & 62.7 & 64.2 & 92.7 & 46.4 \\
GDPO + budget forced & 68.8 & 62.7 & 52.8 & 62.3 & 62.4 & 92.7 & 42.6 \\
Mini-o3 RL recipe~\citep{lai2026mini} & 75.2 & 60.8 & 56.6 & 63.9 & 64.9 & 86.4 & 42.9 \\
Never Give Up~\citep{noukhovitch2026learning} & 74.5 & \textbf{67.9} & 59.4 & 68.0 & 64.9 & 92.2 & 47.3 \\
\rowcolor[HTML]{FAECCC}
\method{} & \textbf{78.7} & 67.5 & \textbf{66.0} & \textbf{70.3} & \textbf{69.1} & \textbf{93.2} & \textbf{53.6} \\
        \midrule
        \multicolumn{8}{@{}l@{}}{\emph{RL comparisons (Gemma4-E4B-it)}} \\
Gemma4-E4B~\citep{team2026gemma} & 26.2 & 15.3 & 5.7 & 16.3 & 16.2 & 58.6 & 27.5 \\
Gemma4-E4B + tools & 35.5 & 32.1 & 20.8 & 30.7 & 22.4 & 70.2 & 28.1 \\
GDPO~\citep{liu2026gdpo} & 57.4 & 48.9 & 27.4 & 46.8 & 40.6 & 79.1 & 32.2 \\
\rowcolor[HTML]{FAECCC}
\method{} & \textbf{61.7} & \textbf{57.1} & \textbf{40.6} & \textbf{55.0} & \textbf{45.0} & \textbf{82.2} & \textbf{35.7} \\
        \bottomrule
    \end{tabular}
    \caption{\textbf{Persistent exploration improves visual search performance across a suite of evaluations and model families.} Accuracy (\%) on VisualProbe, PinpointBench, V*, and O3Bench. RL comparisons use the same training data and initial checkpoint. Bold indicates best performing model within each section.}
    \label{tab:full_results}
\end{fullresultstable}
\noindent\textbf{Baselines.} We compare with the following RL baselines: (1) vanilla GDPO with GSPO policy gradient (GDPO), (2) the GDPO baseline forced to take the max turns, following s1~\citep{muennighoff2025s1} (GDPO + budget forced), (3) the vanilla GDPO setup but with the mini-o3 over-turn masking~\citep{lai2026mini} (Mini-o3 RL recipe), (4) the Never Give Up~\citep{noukhovitch2026learning} method which assigns more rollouts per batch to harder problems by checking if all scores in a group are below 1, and with probability 0.5, generating eight fresh trajectories from the original question. We additionally compare with closed-weight models GPT-6-luna and Sonnet 5 with and without tools, and several recent general reasoning models~\citep{qwen3.5,sarch2026vero,team2026gemma,bai2025qwen3} and specialized models trained for visual search with zooming tools~\citep{lai2026mini,zheng2026deepeyes,gong2026pixeleyes}.

\looseness=-1\noindent\textbf{Training setup.} Our method and all RL comparisons use the same dataset (VisualProbe filtered), same initial checkpoint, group size, training steps, token budgets, and zoom tool. We do a small exploratory sweep of three learning rates per condition and choose the best one after 50 steps. Full RL training details are in the Appendix~\ref{app:training}.

\noindent\textbf{Evaluation suite.} We choose a variety of visual search datasets to test the generalization of the model to challenging and diverse visual search problems. In particular, we evaluate on (1) Visual probe test set~\citep{lai2026mini}, spanning three difficulty levels: easy, medium, and hard, (2) PinpointBench~\citep{gong2026pixeleyes}, comprising ultra-high-resolution images ($\geq$4k), severe distractors, target objects that are extremely
small (typically $\leq$0.07\% of the image), and no spatial question hints (e.g., ``in the bottom-left corner'') or salient anchors (e.g., ``on the huge bridge''), (3) V\rlap{\textsuperscript{*}}~\citep{wu2024v}, a benchmark of high-resolution images from the SA-1B dataset with an average image resolution of 2246$\times$1582, and (4) O3Bench~\citep{li2026insight_o3}, a benchmark composed of map navigation and chart analysis in high resolution images, requiring information from multiple image areas.

\ifdefined\arxivWrapTable\WFclear\fi
\subsection{Persistent exploration overcomes the RL plateau on hard visual search}
\label{sec:results-rq1}

We first investigate how persistent exploration training compares to baselines on learning hard visual search tasks. To determine how well \method{} overcomes the performance plateau observed in Section~\ref{sec:rl-plateaus}, we first compare it with vanilla GDPO on its ability to make progress on novel hard problems. Results are provided in Table~\ref{tab:full_results}.

\looseness=-1On the hard split of VisualProbe, starting from Qwen3-VL-8B-Instruct, \method{} outperforms the GDPO baseline by +16. \method{} additionally improve on easy and medium splits with gains of +4.9 and +5.6, respectively. Compared to the base model, GDPO improves hard +7.5 (42.5$\rightarrow$50.0) while \method{} improves +23.5 (42.5$\rightarrow$66.0), about 3$\times$ in gains over the initial model. These performance improvements hold when evaluating at multiple tool call budgets (Appendix~\ref{app:tool-call-budgets-and-retry-training}; 2 turns 65.2 vs. 63.3, 5 turns 68.9 vs. 63.9, 10 turns 70.3 vs. 62.7). The same trend holds when training on Gemma4-E4B, with +6.6 for GDPO compared to +19.8 for Persistent Exploration. Furthermore, forcing the GDPO model to take additional turns up to the maximum budget provides little benefit (GDPO + budget forcing; Table~\ref{tab:full_results}; 62.3 vs. 62.7 overall), indicating that persistent exploration needs to be learned during training rather than used as a test-time adaptation.

Examining performance on the other visual search benchmarks, \method{} outperforms the GDPO baselines. On O3Bench, which requires collecting information from multiple image regions, \method{} improves over GDPO by +7.2 for Qwen3-VL-8B and +3.5 for Gemma4-E4B, and on PinpointBench by +4.9 and +4.4, respectively. This demonstrates that the gains generalize to tasks outside visual probe.

\begin{findingbox}
Persistent exploration continues to improve on hard problems where vanilla RL plateaus.
\end{findingbox}

We further compare \method{} to general visual reasoners and closed-source models in Table~\ref{tab:full_results}. \method{} outperforms GPT-6-luna + tools on VisualProbe (70.3 vs. 53.8, and 66.0 vs. 47.2 on hard) and on PinpointBench (69.1 vs. 55.0), while closed models remain stronger on O3Bench. \method{} outperforms Qwen3.5-9B + tools on VisualProbe (70.3 vs. 56.7) and Vero-Qwen3i-8B on all four benchmarks, while using the same initial model. Finally, on VisualProbe, PinpointBench, and O3Bench, \method{} improves over previous models specialized for visual search, including Mini-o3-7B, DeepEyes-7B, and PixelEyes-8B.

\begin{findingbox}
Persistent exploration outperforms open general reasoners and specialized visual search models of similar size.
\end{findingbox}

\ifdefined\arxivWrapTable\else
    \begin{table}[!t]
        \ablationTable
    \end{table}
    \begin{figure}[!t]
        \behaviorFigure
    \end{figure}
    \begin{figure*}[!t]
        \qualitativeFigure
    \end{figure*}
\fi

\subsection{Persistent exploration benefits from learning from failure}
\label{sec:results-rq2}

We now investigate the benefits of utilizing the failed context during rollout exploration and for updating the policy. To understand whether retaining the prior failed context after failure is important, we compare to Never Give Up~\citep{noukhovitch2026learning} under the same training data and RL setup as \method{}. Never Give Up instead allocates more rollouts that start from scratch to problems for which the entire group failed to receive accuracy reward. On visual probe, Never Give Up achieves 59.4 on hard and 68.0 overall compared to 66.0 on hard and 70.3 overall for \method{}. The Mini-o3 recipe~\citep{lai2026mini}, which encourages longer exploration by masking trajectories that exceed the turn limit, reaches 56.6 on hard and 63.9 overall. These results suggest that encouraging continued exploration from the failed attempt, rather than restarting from scratch, is more effective for hard visual search problems.

\ifdefined\arxivWrapTable
    \arxivWrapTable[12]{\ablationTable[7.8pt]}
\fi
We further investigate the contribution of learning from the failed attempts by ablating the failed branch loss (w/o $\mathcal{L}_{\text{GSPO}}(b_j)$ in Table~\ref{tab:visual-probe-ablations}). We find that removing the failed branch loss reduces overall accuracy from 70.3\% to 67.8\%, especially hurting performance on easy and medium difficulty tasks (-5.7 and -1.5, respectively). Removing the evaluator loss reduces overall accuracy from 70.3\% to 66.6\% (Table~\ref{tab:visual-probe-ablations}), hurting performance on all VisualProbe splits. Thus, learning from the failed attempts is beneficial, both through a policy gradient loss on the failed branches and an auxiliary loss for supervising retry evaluation.

\begin{findingbox}
Failed attempts provide useful context for exploration and useful signals for training.
\end{findingbox}

\subsection{Persistent exploration replaces early stopping with sustained search}
\label{sec:behavior-analysis}

Accuracy alone does not show how persistent exploration alters the behavior of the model. To complement the evaluation results, we now compare and investigate the reasoning traces across three models: the initial model before training (Qwen3-VL-8B-Instruct), the vanilla GDPO model (GDPO in Table~\ref{tab:full_results}), and the \method{} model. 

\noindent\textbf{Experimental setup.}
To analyze the visual search behavior, we introduce an annotation process using a predefined behavior taxonomy, as detailed in the Appendix~\ref{app:example-trajectories}. To obtain the behavior taxonomy, we analyzed 10 trajectories from each model and identified recurring visual search behaviors. We then use Gemini 3.1 Pro as a judge to annotate each complete reasoning trace of the three models with a binary label indicating the presence or absence of each behavior in the taxonomy. We report the presence rate of each behavior across trajectories in Figure~\ref{fig:behavior-taxonomy}.

\ifdefined\arxivWrapTable
    \noindent\begin{minipage}[t]{0.48\linewidth}\vspace{0pt}
\fi
\noindent\textbf{Findings.}
We find that persistent exploration leads to a higher prevalence of behaviors related to sustained search, involving continuing search exploration before the answer is confidently found. 
This can be observed in systematic coverage behavior, or exploration that systematically inspects multiple plausible target regions, where the presence rate of persistent exploration increases by +28.7 from both the base model and vanilla GDPO. On the other hand, premature stopping behavior, where the model ends its search despite ambiguous evidence, decreases by 19.5 from the base and 14.0 from vanilla GDPO. Persistent exploration models also exhibit more coarse-to-fine zooming, where successive observations become tighter crops around a target (28.8\% base vs. 7.4\% GDPO vs. 77.4\% persistent exploration). In Figure~\ref{fig:qualitative_example}, we show an example of such sustained behavior where the model progressively zooms into the target region until confident of the answer. These findings suggest that persistent exploration encourages the model to examine multiple candidate regions and progressively refine its inspection of promising areas.
\ifdefined\arxivWrapTable
    \end{minipage}\hfill
    \begin{minipage}[t]{0.50\linewidth}\vspace{0pt}
        \captionsetup{type=figure}
        \behaviorFigure
    \end{minipage}\par
\fi

\begin{findingbox}
Persistent exploration searches more systematically than the base model and vanilla RL.
\end{findingbox}

\ifdefined\arxivWrapTable
    \begin{figure*}[!hb]
        \qualitativeFigure
    \end{figure*}
\fi

Given that persistent exploration is trained to identify and retry after failure, we further examine behaviors related to recognizing failed attempts and recovering from unsuccessful search. We first examine error detection, where the model explicitly recognizes a failed crop or a contradiction in its reasoning text. Error detection for persistent exploration increases by +58.7 from the base model and +59.3 from vanilla GDPO. Furthermore, successful replanning, where the model recovers from an unsuccessful search by adjusting its strategy and obtaining useful new observations, increases by +46.6 from the base model and +44.1 from vanilla GDPO. In Figure~\ref{fig:qualitative_example}, we show an example of such failure recognition where the model identifies a related but incorrect target sign and searches another region to find the correct one. Together, these findings suggest that the model more frequently recognizes unsuccessful searches and adjusts its search strategy to obtain new evidence.

\begin{findingbox}
Compared with the base model and vanilla RL, persistent exploration is better at recognizing failure and adapting its search strategy.
\end{findingbox}

%% file: sections/discussion_conclusion_long.tex
\section{Discussion}

We introduced \method{}, a reinforcement learning method to teach models to be persistent after failure for visual search tasks. We guide models to retry after failed attempts during training, and use the failed attempts as context for subsequent exploration and as learning targets for updating the policy. We demonstrated across a suite of challenging visual search benchmarks that learning persistent exploration improves performance over RL baselines and general and specialized visual reasoning models. 

Our findings support the view that learning to recognize and self-correct failed attempts is not an automatic consequence of allocating more compute to on-policy learning. We show that this behavior to recognize failure and correct does not reliably emerge out of standard on-policy RL training (Section~\ref{sec:rl-plateaus}), leading to an inability to progress on harder tasks. This is consistent with \citet{kumar2025training} and \citet{qu2024recursive}, that showed that training on successful responses or offline correction traces does not enable models to reliably self-correct on new tasks. In visual search, models need to evaluate how confident they are based on the evidence they have obtained, and adjust their strategy when the strategy they have deployed so far is not working. Our analysis demonstrates that training persistent exploration amplifies such behaviors, while standard on-policy training does not (Section~\ref{sec:behavior-analysis}).

Persistent exploration training guides models without requiring expert search trajectories, oracle hints, or prefixes of oracle solutions. While prior approaches argue for using oracle guidance when failure occurs~\citep{qu2026pope,zhang2026bread,zhang2026stephint,wang2026beacon}, we show that for visual search, a weaker form of guidance can suffice. Simply preventing an unsuccessful attempt from stopping further exploration while allowing for the model to decide how to persist is itself enough to overcome failure and improve on harder tasks. 


Several areas for future work remain. While our method reduces premature stopping and adapts the number of turns it uses, a natural next step would be to learn more selective persistence. Additionally, we do not investigate context compaction techniques that may enable longer persistence at a lower token cost. Lastly, the work targets visual search as a broad and general domain, but future work may extend the method to other types of language model and vision-language model tools use domains.

\noindent\textbf{Acknowledgements.}
This work was supported by Princeton Research Computing resources, including the Della high-performance computing cluster. We thank Princeton Language and Intelligence (PLI) for their support of this research, as well as Google's TPU Research Cloud (TRC) program for providing additional compute.

%% file: sections/appendix.tex
\appendix
\makeatletter
\renewcommand\thetable{\thesection\@arabic\c@table}
\renewcommand\thefigure{\thesection\@arabic\c@figure}
\renewcommand\thelstlisting{\thesection\@arabic\c@lstlisting}
\@addtoreset{figure}{section}
\@addtoreset{table}{section}
\@addtoreset{lstlisting}{section}
\makeatother
\setcounter{figure}{0}
\setcounter{table}{0}

\begingroup
\IfClassLoadedT{zlab}{%
    \renewcommand{\rmdefault}{ppl}\normalfont
    {\huge\bfseries Appendix}\par
    \smallskip
    \small\setlength{\parskip}{0pt}
}
\hypersetup{linkcolor=black}
\etocsetlocaltop.toc{part}
\etocstandardlines
\etocsettocstyle{}{}
\etocsetnexttocdepth{subsection}
\localtableofcontents
\endgroup
\raggedbottom

\section{Additional Experimental Results}

\subsection{Tool-call budgets and retry training}
\label{app:tool-call-budgets-and-retry-training}

\noindent\textbf{Varying tool call inference budgets.} We vary the max number of tool call budgets from 2, 5, and 10 maximum calls in Figure~\ref{fig:budget}. We include in the comparison \method{}, GDPO, Mini-o3 recipe, and Never Give Up from Table~\ref{tab:full_results}.
\method{} outperforms all baselines tested across tool call budgets. This demonstrates the effectiveness of training persistent exploration even at smaller test time budgets.

\newcommand{\appendixBudgetChart}[1]{%
    \centering
    \includegraphics[width=#1]{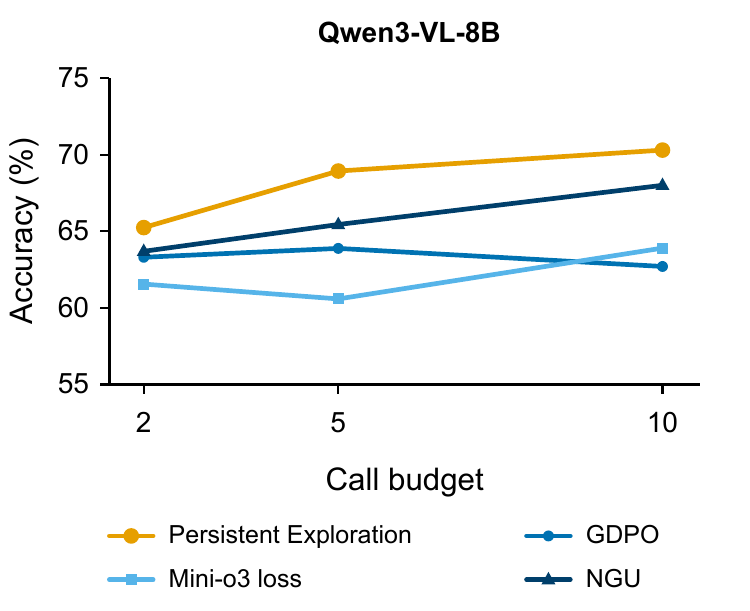}
    \caption{\textbf{\method{} outperforms baselines at varied inference tool call budgets.} We test each model at tool call budgets varying from 2, 5, and 10 maximum calls.}
    \label{fig:budget}
}
\newcommand{\appendixRetryChart}[1]{%
    \centering
    \includegraphics[#1]{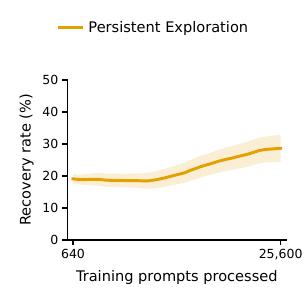}
    \caption{\textbf{Smoothed recovery rate of retrying across training steps}. We use a moving average of 10 steps.}
    \label{fig:retry-recovery}
}
\newcommand{\appendixPrefixText}{\textbf{Unmasking prefixes during training.} Table~\ref{tab:visual-probe-prefix-ablation} compares training with the retry prefix masked or unmasked. Despite GSPO clipping, unmasking the prefix tokens leads to substantially lower performance on visual probe (63.7\%). Thus, masking the prefix is important for performance, as it keeps updates on-policy.}
\newcommand{\appendixPrefixTable}{%
    \centering
    \small
    \setlength{\tabcolsep}{3pt}
    \renewcommand{\arraystretch}{1.12}
    \begin{adjustbox}{max width=\columnwidth}
    \begin{tabular}{@{}lrrrr@{}}
        \toprule
        \multirow{2}{*}{Condition} & \multicolumn{4}{c}{VisualProbe} \\
        \cmidrule(lr){2-5}
        & Easy & Medium & Hard & Overall \\
        \midrule
\method{} & \textbf{78.7} & \textbf{67.5} & \textbf{66.0} & \textbf{70.3} \\
w/ unmasked prefix & 70.9 & 61.9 & 58.5 & 63.7 \\
        \bottomrule
    \end{tabular}
    \end{adjustbox}
    \caption{\textbf{Effect of prefix masking on final accuracy}. We ablate masking the prefix during RL training (w/ unmasked prefix). Unmasking the prefix lowers performance.}
    \label{tab:visual-probe-prefix-ablation}
}
\newcommand{\appendixRetryText}{Figure~\ref{fig:retry-recovery} shows how often Qwen3-VL-8B-Instruct recovers after a forced retry during training.}

\IfClassLoadedTF{zlab}{%
    \begingroup
    \setlength{\parskip}{0pt}
    \captionsetup{skip=3pt,belowskip=0pt}
    \appendixRetryText\par
    \smallskip
    \noindent
    \begin{minipage}[b]{0.53\linewidth}
        \captionsetup{type=figure}
        \appendixBudgetChart{0.88\linewidth}
    \end{minipage}\hfill
    \begin{minipage}[b]{0.44\linewidth}
        \captionsetup{type=figure}
        \appendixRetryChart{width=0.88\linewidth}
    \end{minipage}\par
    \smallskip
    \noindent
    \begin{minipage}[c]{0.46\linewidth}
        \appendixPrefixText
    \end{minipage}\hfill
    \begin{minipage}[c]{0.51\linewidth}
        \captionsetup{type=table}
        \appendixPrefixTable
    \end{minipage}\par
    \endgroup
}{%
    \begin{figure}[!htbp]
        \appendixBudgetChart{0.92\columnwidth}
    \end{figure}
    \appendixPrefixText\par
    \begin{table}[!htbp]
        \appendixPrefixTable
    \end{table}
    \appendixRetryText\par
    \begin{figure}[!htbp]
        \appendixRetryChart{}
    \end{figure}
}



\subsection{Additional qualitative examples}

\newenvironment{appendixExample}
    {\IfClassLoadedTF{zlab}
        {\par\addvspace{\intextsep}\noindent\begin{minipage}{\linewidth}\captionsetup{type=figure}}
        {\begin{figure*}[!t]}}
    {\IfClassLoadedTF{zlab}
        {\end{minipage}\par\nopagebreak[3]\addvspace{\smallskipamount}}
        {\end{figure*}}}

\noindent\textbf{Example 1.} In Figure~\ref{fig:qualitative_hard38}, we show an example asking "What color are the shoes of the person carrying a white shoulder bag?". The \method{} model first searches to identify the white shoulder bag until confident. Then, the model outputs several additional crops to zoom in further until the shoe color is identified. The trace shows how persistent exploration combines attribute-based localization with successive verification to solve this hard problem.

\begin{appendixExample}
    \centering
    \includegraphics[width=\IfClassLoadedTF{zlab}{0.94\linewidth}{\textwidth}]{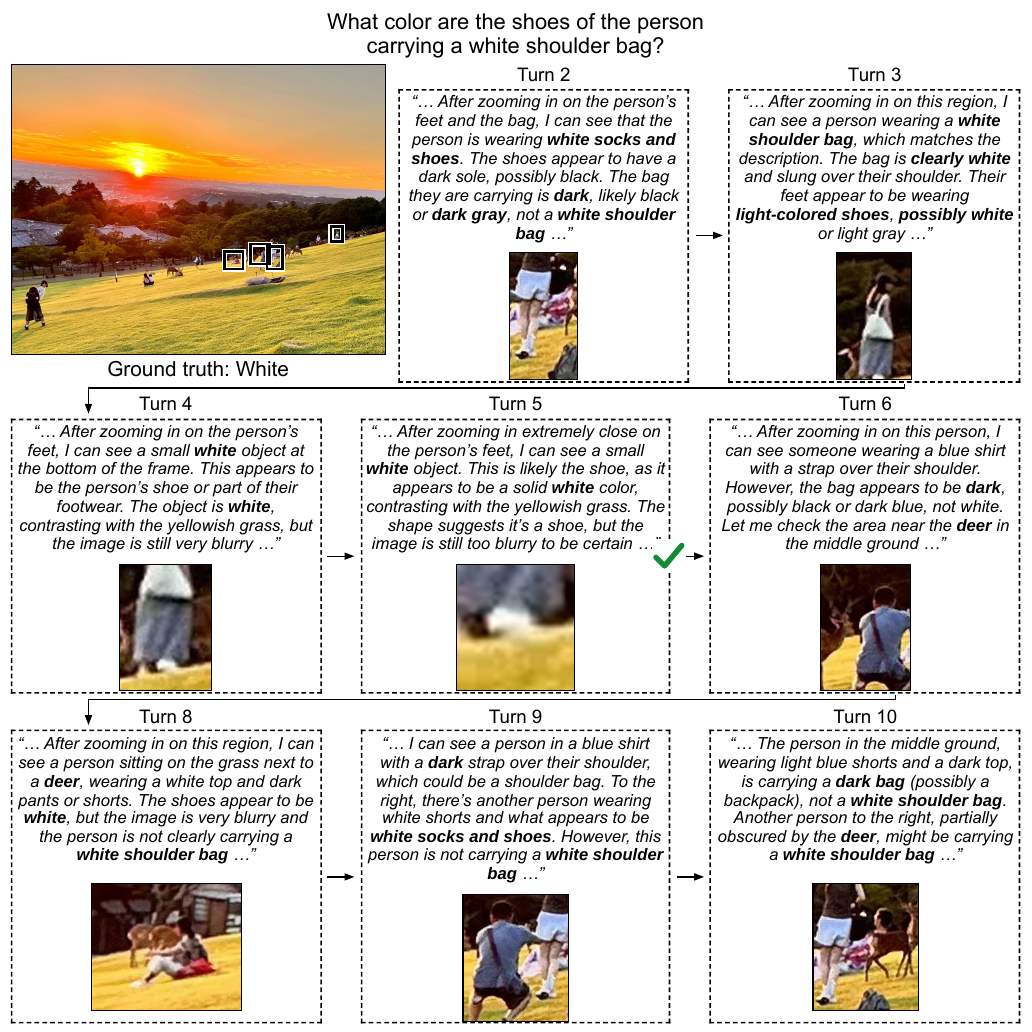}
    \caption{\textbf{Persistent exploration uses a distinctive attribute to locate the target person and zooms in to verify the answer.} \method{} finds a person carrying a white shoulder bag and progressively zooms toward the feet, confirming that the shoes are white. Subsequent views inspect nearby people, including a candidate whose bag is dark rather than white.}
    \label{fig:qualitative_hard38}
\end{appendixExample}

\noindent\textbf{Example 2.} In Figure~\ref{fig:qualitative_medium100}, we show an example of "What is the number on the second row of the blue box on the right side of the picture?". The \method{} detects an initial text area that could be relevant to the task, but upon further inspection and verification, decides that it is the wrong area. It then adapts its search to a different part of the same object, correctly identifying the target region and reading off the correct number. The trace suggests that persistent exploration helps for models to determine an initial incorrect search and keep searching until confident.

\begin{appendixExample}
    \centering
    \includegraphics[width=\textwidth]{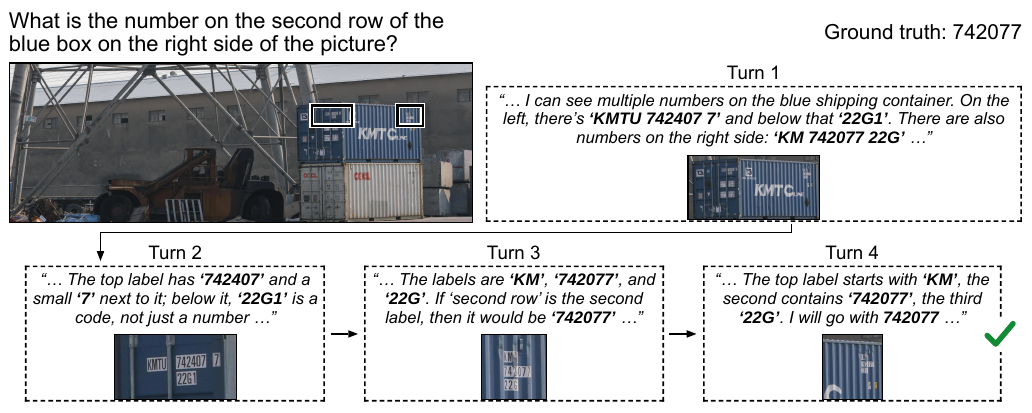}
    \caption{\textbf{Persistent exploration revises an early reading and zooms into the correct region to verify the answer.} \method{} first reads labels on the blue container (``742407'' and ``22G1'') that do not match the queried second-row number on the right. It then searches the right side of the container, where a closer view of the stacked markings confirms the number as 742077.}
    \label{fig:qualitative_medium100}
\end{appendixExample}

\subsection{Trajectory behavior}
\label{app:behavior-analysis}

In this section, we provide further details and expanded results of the behavioral analysis. Table~\ref{tab:zooming-behavior-prevalence} provides individual prevalence rates of each individual behavior. Full behavior definitions are provided in Table~\ref{tab:zooming-behavior-taxonomy}.



\newcommand{\appendixBehaviorTable}{%
\centering
\small
\IfClassLoadedTF{zlab}{\renewcommand{\arraystretch}{1.12}}{\renewcommand{\arraystretch}{1.32}}
\caption{Behavior prevalence rate for the trajectory behavior analysis. Rates include both correct and incorrect trajectories.}
\label{tab:zooming-behavior-prevalence}
\resizebox{\textwidth}{!}{%
\begin{tabular}{@{}llrrr@{}}
\toprule
Dimension & Behavior & Base & GDPO & Persistent exploration \\
\midrule
Search & \texttt{systematic\_coverage} & 22 (4.3\%) & 22 (4.3\%) & 169 (32.9\%) \\
Search & \texttt{redundant\_revisitation} & 26 (5.1\%) & 10 (1.9\%) & 324 (63.2\%) \\
Evidence & \texttt{evidence\_state\_tracking} & 54 (10.5\%) & 41 (8.0\%) & 377 (73.5\%) \\
Evidence & \texttt{independent\_verification} & 5 (1.0\%) & 1 (0.2\%) & 129 (25.1\%) \\
Evidence & \texttt{direct\_unverified\_acceptance} & 43 (8.4\%) & 86 (16.8\%) & 5 (1.0\%) \\
Recovery & \texttt{error\_or\_contradiction\_detection} & 32 (6.2\%) & 29 (5.7\%) & 333 (64.9\%) \\
Recovery & \texttt{successful\_replanning} & 13 (2.5\%) & 26 (5.1\%) & 252 (49.1\%) \\
Recovery & \texttt{backtracking\_and\_correction} & 2 (0.4\%) & 5 (1.0\%) & 72 (14.0\%) \\
Recovery & \texttt{repeated\_failed\_action} & 19 (3.7\%) & 8 (1.6\%) & 232 (45.2\%) \\
Recovery & \texttt{failure\_to\_recover} & 51 (9.9\%) & 15 (2.9\%) & 136 (26.5\%) \\
Stopping & \texttt{evidence\_supported\_stopping} & 318 (62.0\%) & 372 (72.5\%) & 184 (35.9\%) \\
Stopping & \texttt{premature\_stopping} & 150 (29.2\%) & 122 (23.8\%) & 50 (9.7\%) \\
Zooming & \texttt{coarse\_to\_fine\_zooming} & 148 (28.8\%) & 38 (7.4\%) & 397 (77.4\%) \\
Zooming & \texttt{adaptive\_recentering} & 4 (0.8\%) & 4 (0.8\%) & 147 (28.7\%) \\
\bottomrule
\end{tabular}}
}
\IfClassLoadedTF{zlab}{%
    \par\addvspace{\intextsep}
    \noindent\begin{minipage}{\linewidth}
        \captionsetup{type=table}
        \appendixBehaviorTable
    \end{minipage}\par
}{%
    \begin{table*}[!htbp]
        \appendixBehaviorTable
    \end{table*}
}



\noindent\textbf{Findings.}
As noted in the main text, persistent exploration leads to a higher prevalence of behaviors related to sustained search as well as the capability to recognize and recover from failed attempts. We now further investigate the behavior shifts that persistent exploration induces relative to the base model and standard GDPO.

We find that persistent exploration elicits a behavior with better evidence tracking, where the model is more likely to retain and extract information from previous observations. For instance, when prompted to find "a building to the right of the red tower-top building", persistent exploration allows the model to first zoom in on the red tower-top building, then use that information to locate the building to its right. The earlier observation therefore remains useful even after the model moves to another region. This is further supported by the behavior presence rate, where evidence tracking occurs in 73.5\% of trajectories with persistent exploration, compared with 10.5\% for the base model and 8.0\% for standard GDPO.

Additionally, persistent exploration encourages the model to verify its visual interpretations before answering. Verification involves obtaining a distinct view that can confirm an existing hypothesis. We find that the persistent exploration model more often checks a tentative answer through additional observations before committing to it. This is justified by the fact that independent verification occurs in 25.1\% of its trajectories, compared with 1.0\% for the base model and 0.2\% for standard GDPO. Meanwhile, direct unverified acceptance decreases to 1.0\%, compared with 8.4\% and 16.8\%, respectively. Together, these findings suggest that persistent exploration encourages the model to treat an initial interpretation as a hypothesis to check.

However, persistent exploration also introduces redundant search, revealing room to improve exploration efficiency. We find that the model sometimes revisits substantially the same region without obtaining new information or serving a verification purpose. Redundant revisitation occurs in 63.2\% of trajectories with persistent exploration, compared with 5.1\% for the base model and 1.9\% for standard GDPO. Repeated failed actions also increase to 45.2\%, compared with 3.7\% and 1.6\%, respectively. These findings show that continued exploration does not always yield additional evidence, motivating better selection of informative follow-up views.

\paragraph{Persistent exploration produces a closed-loop search profile.}
Error or contradiction detection (64.9\% vs. 6.2\%) and successful replanning (49.1\% vs. 2.5\%) increase relative to the base model, showing that failed observations redirect subsequent search. Coarse-to-fine zooming (77.4\% vs. 28.8\%) and adaptive recentering (28.7\% vs. 0.8\%) become more common. Thus, relative to the base model, persistent exploration learns to retain evidence across views, react to failed observations, and correct its zoom trajectory.



\section{Additional Methods Details}

\subsection{Reward design}
We use four reward terms: an answer accuracy reward, a format adherence reward, a tool efficiency reward, and a crop overlap penalty. They use GDPO weights 0.8, 0.2, 0.15, 0.2, respectively.

\paragraph{Accuracy.} We use a model grader to determine if the predicted answer matches the ground truth. To avoid requiring a separate grading model, we use the rollout policy itself to grade its own extracted answer from text alone, given the question and reference. We assign $r_{\rm acc}=1$ for an equivalent answer and $0.5$ for an answer that identifies the correct target and gives at least half of the required information (e.g., prediction is "white" and ground truth is "red and white"). The judge decodes greedily. If grading fails, we use string matching.

\paragraph{Format.} Each assistant turn must contain a nonempty \texttt{<think>} block followed by one tool call or final answer. For a valid trajectory, $r_{\rm fmt}=1$ if the answer is boxed and $0.5$ if it is nonempty but unboxed. Invalid trajectories receive zero format reward.

\paragraph{Tool efficiency.} We use a concave up penalty from ~\citet{research2026composer} to penalize correct trajectories that take more turns. For $c$ tool calls, let $u=\max(c-1,0)$. The first call is free, and subsequent calls incur a penalty
\[
r_{\rm eff}=-\frac{(1+10u)^{0.15}-1}{10(0.15)}.
\]
We use this component only for $r_{\rm acc}=1$ answers in groups with at least two $r_{\rm acc}=1$ canonical rollouts.

\paragraph{Crop overlap.} We also penalize redundant, highly overlapping, or overly small zoom tool calls. Over valid zoom calls, let $f_{\rm repeat}$ be the fraction with intersection over union of at least $0.90$ with an earlier box on the same image. Let $f_{\rm small}$ be the fraction covering less than $0.0002$ of the image area. A refinement is a smaller, nonrepeated box at least 95\% contained in the immediately preceding valid box, with both boxes on the same image index. We count refinements beyond the first two in each consecutive sequence toward $f_{\rm depth}$. The reward is
\[
r_{\rm overlap}=-f_{\rm repeat}-0.5 f_{\rm small}-0.5 f_{\rm depth}.
\]

\IfClassLoadedF{zlab}{\onecolumn}

\raggedbottom

\subsection{Policy prompts and zoom tool}
\label{app:prompts}

All RL conditions use the same system prompt and tool call template. The system prompt used is displayed in Listing~\ref{lst:policy-system}.

\begingroup
\tcbset{promptstylefmt/.append style={unbreakable, top=3pt, bottom=4pt, before skip=4pt, after skip=5pt}}
\IfClassLoadedT{zlab}{\tcbset{promptstylefmt/.append style={breakable}}}
\newtcolorbox[use counter*=lstlisting]{appendixprompt}[2]{
  promptstylefmt, colframe=black,
  fontupper=\small\fontencoding{T1}\sffamily,
  title={Listing~\thetcbcounter. #1}, title after break={Listing~\thetcbcounter. #1},
  nameref={#1}, label={#2}
}
\lstset{
  style=promptcode, basicstyle=\footnotesize\fontencoding{T1}\ttfamily,
  breakatwhitespace=true, breakautoindent=false, breakindent=0pt,
  aboveskip=2pt, belowskip=2pt,
  literate={’}{{\textquoteright}}1 {“}{{\textquotedblleft}}1 {”}{{\textquotedblright}}1 {–}{{\textendash}}1 {…}{{\textellipsis}}1
}

\begin{appendixprompt}{Policy system prompt}{lst:policy-system}
\setlength{\fboxsep}{1pt}
\fontsize{8.5}{10}\selectfont
You are a helpful, conversational assistant tasked with answering a question about an image.

Every assistant response must contain exactly one non-empty \pcode{<think>} block followed by exactly one action block. The action block must be either \pcode{<tool\_call>} when a tool is needed, or \pcode{<answer>} when the response is final.

Your response must include two parts:

\begin{promptenumerate}
\item \pbold{Reasoning}: A detailed, free-flowing chain of thought enclosed in \pcode{<think>} and \pcode{</think>} tags.
\item \pbold{Action}: Either a tool call enclosed in \pcode{<tool\_call>} and \pcode{</tool\_call>} tags, or a clear final answer enclosed in \pcode{<answer>} and \pcode{</answer>} tags, using \pcode{\textbackslash{}boxed\{\}} notation when the question has a definitive answer.
\end{promptenumerate}

\promptrule

\pbold{Reasoning Instructions}

\begin{promptitemize}
\item The reasoning section must be inside \pcode{<think>} … \pcode{</think>} tags.
\item The reasoning should resemble a stream of consciousness: explore, test hypotheses, backtrack if necessary, reflect, and refine.
\item Let the reasoning flow naturally while progressing toward a conclusion.
\item Use reasoning strategies such as:
\begin{promptitemize}
\item \pbold{Planning} – outline possible approaches before committing.
\item \pbold{Exploration} – consider multiple image regions or interpretations, even unlikely ones.
\item \pbold{Evaluation} – compare alternatives and verify against visual evidence.
\item \pbold{Reflection} – revisit earlier ideas if they may still be viable.
\end{promptitemize}
\item Thoroughly examine and cross-check relevant image regions before narrowing down.
\item If the image is ambiguous, make a reasonable inference based on visual and contextual cues.
\item End the reasoning once you are confident in the conclusion.
\end{promptitemize}

\promptrule

\pbold{Answer Instructions}

\begin{promptitemize}
\item If you need a tool, the action section must be enclosed in \pcode{<tool\_call>} … \pcode{</tool\_call>} tags and contain only the tool-call JSON.
\item The answer section must be enclosed in \pcode{<answer>} … \pcode{</answer>} tags.
\item The \pcode{<answer>} section should stand on its own as a response to the user: it must provide necessary context and justification so that a reader can understand and verify the conclusion without reading \pcode{<think>}.
\begin{promptitemize}
\item Do NOT refer to the \pcode{<think>} section (avoid phrases like “as explained above” or “from the reasoning”).
\end{promptitemize}
\item Boxed result:
\begin{promptitemize}
\item If the question has a definitive, concise answer (a number, word, phrase, or label), include a conversational, natural response followed by exactly one boxed result using LaTeX: \pcode{\textbackslash{}boxed\{final\_result\}}.
\item If the question is open-ended, subjective, or does not yield a concise final result, omit the boxed notation.
\end{promptitemize}
\end{promptitemize}

\promptrule

\pbold{Format Examples}\par\nopagebreak

\noindent\begin{minipage}[t]{0.44\linewidth}
Tool call:

\begin{lstlisting}
<think>
Detailed reasoning goes here...
</think>
<tool_call>
{"name": "<tool-name>", "arguments": {"<argument-name>": "<argument-value>"}}
</tool_call>
\end{lstlisting}
\end{minipage}\hfill
\begin{minipage}[t]{0.54\linewidth}
Final answer:

\begin{lstlisting}
<think>
Detailed reasoning goes here...
</think>
<answer>
Self-contained response goes here...
Following the response, if a concise final result exists, include: \boxed{final_result}. If open-ended or no concise result, respond naturally without \boxed.
</answer>
\end{lstlisting}
\end{minipage}
\end{appendixprompt}

\IfClassLoadedF{zlab}{\clearpage}
\tcbset{promptstylefmt/.append style={breakable, top=7pt, bottom=7pt, before skip=8pt, after skip=8pt}}
\begin{appendixprompt}{Training zoom schema}{lst:train-zoom}
\begin{lstlisting}
{
  "type": "function",
  "function": {
    "name": "image_zoom_in_tool",
    "description": "Zoom in on a specific region of an image by cropping it based on a bounding box (bbox) and an optional object label",
    "parameters": {
      "type": "object",
      "properties": {
        "bbox_2d": {"type": "array",
          "description": "The bounding box of the region to zoom in, as [x1, y1, x2, y2], where (x1, y1) is the top-left corner and (x2, y2) is the bottom-right corner"},
        "label": {"type": "string",
          "description": "The name or label of the object in the specified bounding box"},
        "img_idx": {"type": "number",
          "description": "The index of the zoomed-in image (starting from 0)"}
      },
      "required": ["bbox_2d", "label", "img_idx"]
    }
  }
}
\end{lstlisting}
\end{appendixprompt}

At evaluation, the \texttt{bbox\_2d} property also specifies \verb|"items": {"type": "number"}|, \verb|"minItems": 4| and \verb|"maxItems": 4|. Image index~0 refers to the original image. Crops receive successive indices. Box coordinates range from 0 to 1,000 in the indexed view. We crop from the original-resolution source when available.

\begin{appendixprompt}{Zoom tool response text}{lst:crop-response}
\begin{lstlisting}
Zoomed in on the image to the region {bbox_2d} with label {label}.
\end{lstlisting}
\end{appendixprompt}

\begin{appendixprompt}{Retry prefill}{lst:retry-prefill}
Let’s see. I still don’t have decisive evidence, so I should inspect a substantially different region. If one area looks especially promising, I can examine it more closely, but I should look elsewhere once further zooming stops revealing new evidence.
\end{appendixprompt}

\begin{appendixprompt}{Final-answer prefill}{lst:final-prefill}
This is the final turn, so I need to answer from the evidence gathered so far.
\end{appendixprompt}

In Listing~\ref{lst:accept-retry}, \texttt{\{candidate\}} contains the reasoning and answer from the candidate assistant turn.

\begin{appendixprompt}{Accept/retry transcript}{lst:accept-retry}
\begin{lstlisting}
<|im_start|>assistant
{candidate}<|im_end|>
<|im_start|>user
Assuming another tool round is available, should the preceding candidate be accepted or retried? In the <answer> block, output only "A" or "R".<|im_end|>
<|im_start|>assistant
<think>
I need to classify whether the preceding candidate should be accepted or retried.
</think>
<answer>
\end{lstlisting}
\end{appendixprompt}

\endgroup

\IfClassLoadedTF{zlab}{\Needspace{6\baselineskip}}{\twocolumn}

\section{Additional Training Details}
\label{app:training}

\subsection{Data and RL hyperparameters}

\newenvironment{appendixSettings}[1]{%
    \IfClassLoadedT{zlab}{\noindent\begin{minipage}[c]{#1}\setlength{\parskip}{0.5\baselineskip}}%
}{%
    \IfClassLoadedTF{zlab}{\end{minipage}\par\addvspace{\medskipamount}}{\end{table}}%
}
\newcommand{\appendixSettingsTable}[1]{%
    \IfClassLoadedTF{zlab}{%
        \end{minipage}\hfill
        \begin{minipage}[c]{#1}\captionsetup{type=table}
    }{\begin{table}[!ht]}%
}
\newcommand{\appendixSettingsRows}[2]{%
    \IfClassLoadedTF{zlab}{%
        \begin{tabularx}{\linewidth}{@{}l@{\hspace{0.8em}}>{\raggedright\arraybackslash}X@{}}
            #2
        \end{tabularx}
    }{%
        \begin{tabular}{@{}l@{\hspace{0.8em}}p{#1\columnwidth}@{}}
            #2
        \end{tabular}
    }%
}

\begin{appendixSettings}{0.345\linewidth}
\noindent\textbf{Training dataset.} For curating our training dataset, we filter the original VisualProbe to keep only hard samples that require zooming to answer. We run Vero-Qwen3i-8B~\citep{sarch2026vero} (text-only, no tools) on each question 16 times. Of the 5,330 questions in the original train set, we keep 2,042 that receive no correct answer in any attempt.

\noindent\textbf{Hyperparameters.} We provide complete training hyperparameters in Table~\ref{tab:appendix-training}.

\appendixSettingsTable{0.62\linewidth}
\centering
\caption{Training setup.}
\label{tab:appendix-training}
\small
\renewcommand{\arraystretch}{1.1}
\appendixSettingsRows{0.65}{%
\toprule
Setting & Value \\
\midrule
Batch & 128 prompts $\times$ 8 rollouts, 64 per minibatch \\
Optimizer & AdamW, $\beta=(0.9,0.999)$, weight decay 0.01, gradient norm clip 1.0 \\
LR sweep & $5\times10^{-6}$, $1\times10^{-5}$, $5\times10^{-5}$ \\
Policy update & GDPO advantages, GSPO loss, ratio clipped to $[0.9997,1.0004]$, no KL penalty \\
Sampling & $T=1$, top-$p=1$ \\
Token limits & 16,384 input, 24,576 response tokens, 40,960 total, 4,096 per assistant turn \\
Turn limits & 11 assistant turns, up to 10 tool responses, one zoom per turn \\
Images & 10M input pixels, crops resized to min 518,400, max 1,048,576 pixels \\
Training & 8 GPUs, BF16 with FSDP2 \\
\bottomrule
}
\end{appendixSettings}

\subsection{Baseline training}

\paragraph{Mini-o3 Baseline.}
We compare to the Mini-o3~\cite{lai2026mini} overturn masking as a baseline. We train the baseline on the same dataset as our method and RL setup. Instead of forcing an answer at max turns, rollouts that exceed either the turn or context-length limit are excluded from the actor update.

\paragraph{Never Give Up Baseline.}
We adapt the Never Give Up~\citep{noukhovitch2026learning} prompt resampling to our GDPO / GSPO setup. For each prompt, we sample 8 complete trajectories and retain groups with varying accuracy rewards. We discard groups that all score correct and retry another 8 trajectories from scratch with probability 0.5 for groups with all incorrect scores. If a retry improves the accuracy reward, we pool all attempts for that prompt in the same policy update.





\section{Additional Evaluation Details}
\label{app:evaluation}

\subsection{Decoding and scoring}

\begin{appendixSettings}{0.515\linewidth}
\noindent\textbf{Evaluation scoring.} We take the text after the last tool response and extract the contents of the \texttt{<answer>} block. If the answer contains a complete \verb|\boxed{...}|, we use the contents of the box. We give the extracted answer and reference to Qwen3.5-27B along with the question and ask whether the answers are equivalent.

\noindent\textbf{Evaluation settings.}
We provide the complete evaluation setup and hyperparameters in Table~\ref{tab:external-decoding}. We use these settings for all baselines unless otherwise indicated.

\appendixSettingsTable{0.45\linewidth}
\centering
\small
\caption{Decoding settings for external evaluation.}
\label{tab:external-decoding}
\renewcommand{\arraystretch}{1.1}
\appendixSettingsRows{0.66}{%
\toprule
Setting & Value \\
\midrule
Sampling & $T=0.7$, top-$p=0.8$, top-$k=20$ \\
Penalties & Repetition 1.0, presence 0.0 \\
Token limits & 40,960 context length \\
Image limit & 10M max pixels \\
Zoom & Hermes parser, 10 turns/tool call budget \\
\bottomrule
}
\end{appendixSettings}

\noindent\textbf{Final answer forcing.} After the last allowed tool round, we disable tool calls and force an answer. If the forced answer is not formatted correctly, we repeat the request once with the reasoning block already closed.

\noindent\textbf{Inference retry evaluator.} For \method{}, we check each candidate answer, including a valid forced final answer. We append the candidate and the accept/retry prompt to the transcript and decode one token from $\{\texttt{A},\texttt{R}\}$ with $T=0$ and top-$p=1$. If $P(\texttt{R})>0.5$ and another tool round and image are available, we return to the history before the candidate and force a zoom using the retry prefix. If the model recommends a retry but cannot continue, we keep the candidate and record a forced accept. We also keep the candidate if the accept/retry request fails.

\noindent\textbf{GDPO with budget forcing.}
We include an inference time budget forcing experiment inspired by s1~\citep{muennighoff2025s1}. At each turn, we constrain the output after each reasoning block to force a tool call token (\texttt{<tool\_call>}). At the final turn, we force an answer by constraining decoding to output an answer tag (\texttt{<answer>}) after the reasoning block. Decoding setup and evaluation protocols are otherwise unchanged.

\subsection{Trajectory annotation}
\label{app:example-trajectories}

\paragraph{Annotation protocol.}
To construct the taxonomy, we provided a frontier model with 10 trajectories from each model and asked it to identify recurring visual-search behaviors. Subsequently, we manually inspected the proposed categories alongside the traces and merged overlapping behaviors, yielding 14 behaviors as given in Table~\ref{tab:zooming-behavior-taxonomy}. After obtaining the final taxonomy, we provided the behavior definitions to Gemini 3.1 Pro, which receives the question, reference answer, prediction, original image, complete reasoning trace, and all reconstructed crops. It then assigns a binary presence label to each complete trajectory for each behavior. We obtain one judgment per trajectory at temperature 0.1 with high reasoning effort. The judge is instructed to cite supporting events and explain positive labels so that the presence of a behavior must be supported by the trace rather than inferred from correctness alone. Finally, we calculate the behavior presence rate, which is the number of positive labels divided by the total number of trajectories.

\begin{table*}[!htbp]
\centering
\caption{Complete 14-label zooming trajectory taxonomy. Each label records
whether a behavior occurs anywhere in the complete trajectory; multiple labels may apply.}
\label{tab:zooming-behavior-taxonomy}
\small
\setlength{\tabcolsep}{3pt}
\begin{tabular}{@{}p{0.10\textwidth}>{\raggedright\arraybackslash}p{0.26\textwidth}p{\dimexpr0.64\textwidth-4\tabcolsep\relax}@{}}
\toprule
Dimension & Behavior & Definition \\
\midrule
Search & \texttt{systematic\_\allowbreak{}coverage} & Searches several regions that could contain the target in an organized way, such as scanning from left to right. \\
Search & \texttt{redundant\_\allowbreak{}revisitation} & Returns to substantially the same region without changing the view in a useful way, obtaining new information, or checking a specific hypothesis. \\
Evidence & \texttt{evidence\_\allowbreak{}state\_\allowbreak{}tracking} & Correctly reuses information from an earlier view when interpreting a later view or deciding where to search next. \\
Evidence & \texttt{independent\_\allowbreak{}verification} & Obtains a different view that can confirm or challenge a tentative interpretation. \\
Evidence & \texttt{direct\_\allowbreak{}unverified\_\allowbreak{}acceptance} & Accepts the first plausible observation despite a specific unresolved ambiguity that needs another check. \\
Recovery & \texttt{error\_\allowbreak{}or\_\allowbreak{}contradiction\_\allowbreak{}detection} & Explicitly points out a problem with its crop or interpretation, such as missing the target or finding conflicting evidence. \\
Recovery & \texttt{successful\_\allowbreak{}replanning} & After an unsuccessful observation, changes where or how it searches and obtains new, useful evidence. The final answer need not be correct. \\
Recovery & \texttt{backtracking\_\allowbreak{}and\_\allowbreak{}correction} & Explicitly changes an earlier visual interpretation in response to new evidence, such as recognizing that a previously selected object was the wrong target. \\
Recovery & \texttt{repeated\_\allowbreak{}failed\_\allowbreak{}action} & Repeats a crop that has already failed to provide useful evidence, without adjusting it to address the problem or explaining what new information it could provide. \\
Recovery & \texttt{failure\_\allowbreak{}to\_\allowbreak{}recover} & An observation fails to provide needed evidence, but the model answers without obtaining a useful correction even though a practical corrective action is available. \\
Stopping & \texttt{evidence\_\allowbreak{}supported\_\allowbreak{}stopping} & Answers promptly once the visual details needed to support the answer are clear and no relevant ambiguity remains. \\
Stopping & \texttt{premature\_\allowbreak{}stopping} & Answers while a specific evidence gap remains apparent in the trace and a feasible additional crop could resolve it. \\
Zoom control & \texttt{coarse\_\allowbreak{}to\_\allowbreak{}fine\_\allowbreak{}zooming} & Starts with a broad region and progressively zooms in on the target, using what it sees at each step to choose the next crop. \\
Zoom control & \texttt{adaptive\_\allowbreak{}recentering} & Moves or expands a crop after noticing that the target is cut off, near the edge, or otherwise difficult to inspect in the current view. \\
\bottomrule
\end{tabular}
\end{table*}


\noindent\textbf{Failure modes.} For Figure~\ref{fig:base-model-error-breakdown}, GPT-6.1-xhigh assigns each incorrect base-model trajectory one label. A \emph{give-up} response is one where the model responds saying that the target is absent or the answer cannot be determined. A \emph{premature answer} gives an unhedged wrong answer after one zoom. An \emph{unresolved search} uses multiple zooms but still fails to find the answer. The remaining failures that cannot be classified into one of three above fall under \emph{other errors}.